\documentclass[acmsmall]{acmart}

\AtBeginDocument{%
  \providecommand\BibTeX{{%
    \normalfont B\kern-0.5em{\scshape i\kern-0.25em b}\kern-0.8em\TeX}}}

\acmJournal{JACM}
\acmVolume{37} 
\acmNumber{4}
\acmArticle{1}
\acmMonth{8}

\usepackage{amsmath}
\usepackage{amsfonts}
\usepackage{algorithm} 
\usepackage[noend]{algpseudocode}

\usepackage{textcomp}
\usepackage{xcolor}
\usepackage{hyperref}
\usepackage{subcaption}
\usepackage{graphicx} 
\usepackage{float}

\usepackage{algorithm}
\usepackage{algpseudocode}

\usepackage{relsize} 
\usepackage{array} 
\usepackage{bbm}
\usepackage{setspace}

\usepackage{booktabs}
\usepackage{multirow}
\usepackage{siunitx}
\usepackage{boondox-cal}
\usepackage[bottom]{footmisc}
\usepackage{tikz}
\usetikzlibrary{positioning}
\usepackage{pgfplots}
\usepackage{microtype}

\usepackage[belowskip=2pt,aboveskip=0pt]{caption}

\begin{document}

\setcopyright{acmcopyright}
\acmJournal{TIST}
\acmYear{2020} \acmVolume{1} \acmNumber{1} \acmArticle{1} \acmMonth{1} \acmPrice{15.00}\acmDOI{10.1145/3442390}


\title{Predicting Attributes of Nodes Using Network Structure}

\author{Sarwan Ali}
\email{16030030@lums.edu.pk}
\affiliation{%
	\institution{Lahore University of Management Sciences}
	\city{Lahore}
	\country{Pakistan}
}

\author{Muhammad Haroon Shakeel}
\email{15030040@lums.edu.pk}
\affiliation{%
	\institution{Lahore University of Management Sciences}
	\city{Lahore}
	\country{Pakistan}
}

\author{Imdadullah Khan}
\email{imdad.khan@lums.edu.pk}
\affiliation{%
	\institution{Lahore University of Management Sciences}
	\city{Lahore}
	\country{Pakistan}
}

\author{Safiullah Faizullah}
\email{safi@iu.edu.sa}
\affiliation{%
	\institution{Islamic University}
	\city{Madinah}
	\country{KSA}
}

\author{Muhammad Asad Khan}
\email{asadkhan@hu.edu.pk}
\affiliation{%
	\institution{Hazara University}
	\city{Mansehra}
	\country{Pakistan}
}
 
\renewcommand{\shortauthors}{Ali, et al.}
\begin{abstract}
In many graphs such as social networks, nodes have associated attributes representing their behavior. Predicting node attributes in such graphs is an important task with applications in many domains like recommendation systems, privacy preservation, and targeted advertisement. Attributes values can be predicted by treating each node as a data point described by attributes and employing classification/regression algorithms. However, in social networks, there is complex interdependence between node attributes and pairwise interaction. For instance, attributes of nodes are influenced by their neighbors (social influence), and neighborhoods (friendships) between nodes are established based on pairwise (dis)similarity between their attributes (social selection). In this paper, we establish that information in network topology is extremely useful in determining node attributes. In particular, we use self and cross proclivity measures (quantitative measures of how much a node attribute depends on the same and other attributes of its neighbors) to predict node attributes. We propose a feature map to represent a node with respect to a specific attribute $a$, using all attributes of its $h$-hop neighbors. Different classifiers are then leaned on these feature vectors to predict the value of attribute $a$. We perform extensive experimentation on ten real-world datasets and show that the proposed method significantly outperforms known approaches in terms of prediction accuracy.
\end{abstract}


\maketitle

\section{Introduction}
In many social and collaboration networks, nodes have additional information
(attributes) associated with them. Such {\em attributed graphs} are becoming increasingly common \cite{goyal2018graph,meng2018dissimilarity}. The attributes could be gender or age, etc. of people in social networks \cite{Prone,liao2018attributed}, research interests of authors in collaboration and citation networks \cite{ye2017attributed}, and structural or functional properties of the proteins in biological networks \cite{hamilton2017inductive,kovacs2019network}.

The values of attributes reflect the characteristics, behavior, and preferences of the entities represented by the nodes. Knowledge of these attributes lead to enhanced recommendation systems \cite{Recurrent2018che,lopes2019graph}, improved community detection \cite{sun2019community,yang2013community,pool2014description}, and robust privacy-preserving mechanism \cite{soria2018differentially,ToJoinOrNotToJoin}. Node attributes can be utilized for graph summarization \cite{Wu2014AttributedSummarization,Beg2018Scalable}. They also play a key role in improving the performance of disease outbreak detection \cite{cook2011assessing,Abbas2017SRE,Ahmad2016AusDM,ahmad2017spectral,Tariq2017Scalable,AHMAD2020Combinatorial} and early depression identification \cite{eric2014characterizing,colombo2016analysing}. In protein-protein interaction networks, attributes of proteins have been used in conjunction with network structure for protein classifications \cite{hamilton2017inductive, hou2019representation}. Additional information about nodes (attributes values) helps design the network embeddings and graph classification algorithms efficiently \cite{li2019adversarial}.

In many practical scenarios, the values of all attributes are not known for all nodes, which limits the usefulness of the networks. An attribute of nodes can be inferred by considering each node as a feature vector (of dimensions equal to the number of attributes). Standard classification/regression algorithms are then employed on these feature vectors for attribute prediction~\cite{alam2016improving, jia2018attriguard, su2011using}. However, this approach does not use the rich information, which is available in the form of interconnection among the nodes.

\begin{figure}[!h]
    \centering
    \includegraphics[width=.93\textwidth]
    {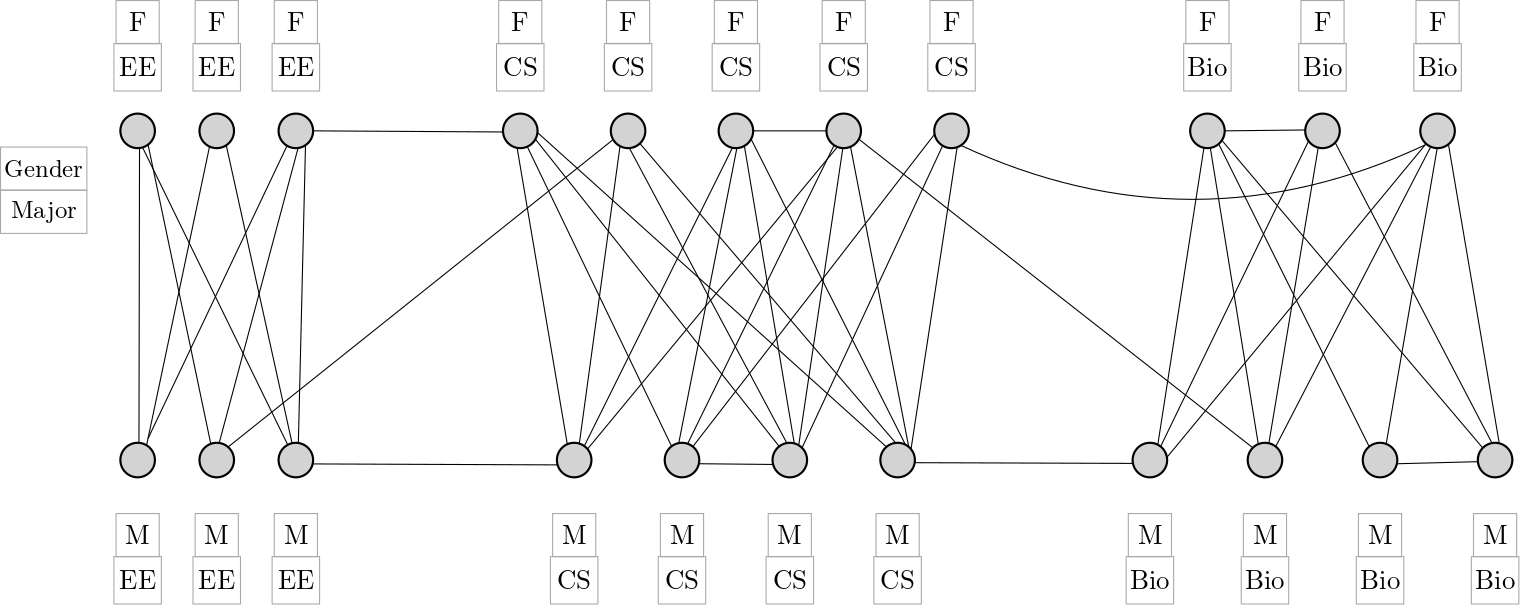}
    \caption{In this hypothetical friendship network of students, each node has attributes gender and academic major. The ``gender" attribute in this network exhibits heterophily (students tend to be friends with those having opposite gender), while the ``major" attribute is homophilic (relatively more edges between students with common majors).
    }
    \label{proclivity_diagram}
\end{figure}

It is well known in the sociology literature that there are two kinds of interdependence between the structure of a network and attributes of nodes, namely {\em social selection} and {\em social influence}~\cite{SocialSelectionAndPeerInfluenceInAnOnlineSocialNetwork}. Social selection refers to the phenomenon where the similarity between nodes attributes leads to edges between them, while social influence states that edges between nodes lead to the similarity between them. Moreover, node attributes in networks exhibit the properties of {\em homophily} or {\em heterophily} \cite{Prone}. Homophily (heterophily) refers to the tendency of nodes with certain values of an attribute to connect with other nodes having the same (different) values for that attribute. Figure \ref{proclivity_diagram} depicts an example network with one homophilic and one heterophilic attribute. The degree of homophily/heterophily is referred to as {\em self proclivity}. The notion of proclivity is extended in \cite{Prone} to define {\em cross proclivity}, which is a measure of correlation among values of different attributes. 

The information in network structure has been utilized for attribute prediction \cite{kim2012latent,yang2018multi, liao2018attributed}, node classification \cite{duran2017learning,hamilton2017inductive,hou2019representation,liao2018attributed}, and other problems in graphs analysis \cite{khot2015gradient,neville2007relational,hassan2020estimating}. Similarly, the effect of homophily has been well explored for node attribute prediction \cite{liao2016slr}. However, the cross proclivity and dependence of a node attribute on more than one attribute of its neighbors has not been utilized to its full extent.

In this paper, we propose {\em Neighborhood based Feature Vector Representation (\textsc{n-fvr})} to predict node attributes in a network. We devise a feature map for nodes with respect to an attribute. These feature vectors are based on the attribute values of the node and those of the {\em ``nearby''} nodes (neighbors, neighbors-of-neighbors, and so on). Standard classification/regression models are learned from the feature vectors to predict attribute values. Following are the main features of \textsc{n-fvr}.
\begin{enumerate}
    \item \textsc{n-fvr} works for both nominal and numerical attributes and is more generally applicable.
    \item \textsc{n-fvr} uses all attributes of nearby nodes to predict an attribute value for a node. Thus it captures the inherent self and cross proclivities in the network.
    \item Feature vector representation of nodes is based on multiple hops neighborhood capturing long term dependencies between attributes that improve predictive performance. 
    \item Extensive evaluation of \textsc{n-fvr} on ten benchmark datasets demonstrates that it significantly outperforms the best-known methods for this task. \textsc{n-fvr} achieved up to $83.64\%$ improvement as compared to \textsc{nns}, up to $21.3\%$ improvement from node2vec, up to $22.2\%$ improvement from DeepWalk, up to $20.1\%$ improvement from \textsc{line}, up to $26\%$ improvement from GraRep, and up to $15.4\%$ improvement from \textsc{mne} in terms of accuracy. 
\end{enumerate} 

The rest of the paper is organized as follows. In Section \ref{relatedwork}, we discuss related work to our problem. We provide detail of our feature map generation in Section \ref{proposedalgorithm}. In Section \ref{experiments}, we describe the implementation details of comparisons methods and different classification/regression algorithms. Section \ref{Experimental_Evaluation} presents the results of our method and comparison. We conclude the paper in Section \ref{conclusion} and discuss some future directions.

\section{Related Work}\label{relatedwork}  

With increasing volume and decreasing veracity of network data with node attributes, the problem of data imputation has attracted significant attention from researchers \cite{sergey2012icdm, neil2014tist, horne2019robust, yangjane}. Both supervised \cite{alam2016improving, su2011using} and unsupervised \cite{zhang2011System} machine learning methods have been used to predict node attributes. These methods consider each node as a data point described by its attributes. For online social networks, additional features extracted from the contents shared by users have been used to predict node attributes. A method based on linguistic features (verbs, pronouns, articles, and prepositions) of social media content is proposed in \cite{hosseini2016recognizing} to determine users' gender. Similarly, in \cite{PrivateTraitsAndAttributes}, the model uses ``likes" of people on Facebook to find binary attributes (single vs. in-relationship, smoker vs. non-smoker, etc.) of users.

These approaches, however, do not utilize the rich information in the topology of the network \cite{Prone,SocialSelectionAndPeerInfluenceInAnOnlineSocialNetwork,liao2018attributed}. Graph clustering based methods are proposed in \cite{MissingDataAnalysesHybrid, perozzi2018discovering} to predict characteristics of nodes based on their communities.
Exploiting homophily of attributes  \cite{ToJoinOrNotToJoin,al2012homophily} uses friendship links and group information in social media to predict users' attributes. However, the underlying assumption of having more homogeneous communities in social networks restricts the applicability as well as limits the predictive accuracy of these approaches. A two-phase (clustering-semantic similarity) approach to predict attributes values of nodes in a network is proposed in \cite{abid2017twoPhase}.

In the representation learning approach, nodes are first mapped to feature vectors (node embedding) using their attributes and network connectivity \cite{cui2018survey,zhang2018network,yangjane,zhu2019pcane}. A classifier is then learned on these vectors to predict node attributes.  
The DeepWalk approach in \cite{perozzi2014deepwalk} employs skip-gram (a word representation model) to learn node representations in the graph. In \cite{tang2015line}, large-scale information network embedding (\textsc{line}) is proposed to learn  representations by preserving the first- and second-order proximities of nodes. 
GraRep \cite{grover2016node2vec} uses global structural information for learning low dimensional node embedding for weighted graphs. One of the most common node embedding approaches is node2vec \cite{cao2015grarep}, which learns continuous feature vectors for nodes using a flexible biased random walk that can explore neighborhoods in both breadth-first search and depth-first search fashion. The latent multi-group membership graph model (\textsc{lmmg})\cite{kim2012latent} summarizes the network structure, predict edges between nodes, and estimate attributes values. 
Several studies have proposed methods to incorporate different facets/structures of nodes into representation learning \cite{yang2018multi,liu2019single}. These multi-facets based models aim to separately capture different characteristics of the network rather than designing single feature vector for all characteristics.
In \cite{duran2017learning}, an unsupervised approach called embedding propagation (\textsc{em}) for graph-structured data is devised. A framework, {\em Joint Adversarial Network Embedding} (JANE) is proposed  in \cite{yangjane} to efficiently capture semantic variations in data distribution.
Yang {\em et al.} \cite{yang20183} analyzes correlations between topological and non-topological features for a network embedding approach in which nodes, communities, and topics (attributes) are mapped into one embedding space.

The representation learning approach has yielded great success in capturing complex relationships across various disciplines such as biological networks and social media \cite{cui2018survey,zhang2018network}. 
In this approach, the dataset is mapped to a fixed-dimensional vector space, and machine learning methods are employed on these feature vectors for classification. In a highly successful method, these feature vectors are based on counts of various substructures in the objects. A kernel function to estimate the pairwise similarity between objects is then defined and used for classification. This approach has been used for classifying images \cite{Bo_ImageKernel}, sequences \cite{Farhan_SequenceKernel,Kuksa_SequenceKernel}, and graphs \cite{kondor2016multiscale, shervashidze2011weisfeiler,zhu2019pcane}. In the descriptors learning approach, objects are mapped to low dimensional vectors of features extracted from objects with the goal to map similar objects closely in the Euclidean space. Many descriptors have been proposed for sequences \cite{atzori2014electromyography,ullah2020effect} electricity consumptions \cite{ali2019short,Ali2020ShortTerm} and graphs \cite{sge,tsitsulin2018netlsd,verma2017hunt,shin2018tri,hassan2020estimating}. More recent but computationally expensive methods employ deep networks together with domain specific techniques for embedding nodes \cite{duran2017learning}, graphs \cite{xu2018powerful, morris2019weisfeiler} and texts \cite{Shakeel2019MultiBilingual,Shakeel2020Multi,Shakeel2020LanguageIndependent}.

\section{Proposed Approach}\label{proposedalgorithm}

In this section, we formulate the problem of node representation and attribute prediction and provide detail of the proposed method, {\em Neighborhood based Feature Vector Representation (\textsc{n-fvr})} of nodes in attributed graphs. 


\subsection{ Notation and Problem Formulation}
Given an undirected attributed graph $G = (V,E,A)$, where $V$ is the set of nodes, $E$ is the set of edges, $E  \subset {V \choose 2}$, and $A$ = $\{a_{1}, a_{2}, \ldots , a_{t} \}$ is the set of $t$ node attributes, an integer $h$, and a target attribute $a_i \in A$, our goal is to predict the value of $a_i$ for nodes. We propose a vector representation $R^{h}_{a_{i}}(v)$ for each vertex $v \in V$, based on the attributes of $v$ and values of all attributes of nodes that are {\em `close by'} to $v$. The notion of closeness is determined by the value of parameter $h$. These vectors are then passed to a classification or regression model with the target attribute ($a_i$) as class label, to predict the value of $a_i$ for nodes.

Attributes can be nominal, ordinal or numerical. We assume all numerical attributes take discrete values (this can be achieved by appropriate discretization). Each attribute $a_i$ takes exactly one value from the set $L_{i}$, where $L_{i} = \{ l_{i_{1}}, l_{i_{2}}, \ldots , l_{i_{n_{i}}} \}$ and $n_{i} = \vert L_{i} \vert$. We assume that for every $i$, $L_{i}$ contains a special symbol for missing value. For instance, if $a_i$ is the gender attribute, then $L_{i} = \{ M, F, \square\}$, where $\square$ represents missing value. We consider the attribute $a_{i}$ as a function such that $a_{i}: V \rightarrow L_{i}$. The set $A$ maps each node $v\in V$ to a $t$-dimensional vector, whose $i^{th}$ coordinate is $a_i(v)$. Hence, $A$ is a vector valued function given by:
\begin{equation}\label{attrib_definition} 
    A: V \rightarrow L_1 \times L_2 \times \ldots \times L_t
\end{equation}

Let $N(v) = \{u\in V: (u,v)\in E\}$ be the set of neighbors of $v$ and let $deg(v) = |N(v)|$ be the degree of $v$. We describe the notion of nodes that are ``close to" $v$. For an integer $h>0$, denote by $N^{h}(v)$ the set of nodes that are $h$ hops away from $v$. Formally:
\begin{equation}\label{h_hop_neighborhood_distance}
    N^{h}(v) = \{u \in V, d(u,v) = h \},
\end{equation}
where $d(u,v)$ is the distance between node $u$ and $v$ (hop-length of the shortest path between $u$ and $v$). Note that $N^{0}(v) = \{v\}$ and $N^{1}(v) = N(v)$.

\subsection{Aggregating attributes values of sets of nodes}
Note that attributes are defined for individual nodes (see Equation \eqref{attrib_definition}) and not for sets of nodes. We therefore, extend the notion of attributes to sets of nodes. For a set $S \subseteq V$, we define $A_{j}$ to be the vector valued function to determine the ``value" of an attribute $a_{j}$ of the set $S$. 
Recall that $a_{j}: V \rightarrow L_{j} = \{ l_{j_{1}}, l_{j_{2}}, \ldots l_{j_{n_{j}}} \}$. For a set $S$, the function $A_j$ returns a $|L_j|=n_j$ dimensional real vectors, that is essentially the (discrete) distribution of values of $a_{j}$ in the set $S$. More formally, for $S\subseteq V$, the function $A_{j}: 2^{V} \rightarrow \mathbb{R}^{n_{j}}$ is defined as follows:
\begin{equation} \label{single_attribute_equation_SAQ}
    \left( A_{j}(S)\left[k\right] = \frac{1}{\vert S \vert} (|\{ y \in S : a_{j}(y) = l_{j_k} \}|)\right)_{1 \leq k \leq n_{j}} \text{ for } l_{j_k} \in L_j, and \ S \neq \emptyset,
\end{equation}
where $X[k]$ is the $k^{th}$ coordinate of the vector $X$.

\subsection{Pairwise interconnection between attributes: self and cross proclivity}

To find the interconnection between two attributes $a_{i}$ and $a_{j}$, we analyze the global connectivity structure of input graph with respect to attributes $a_{i}$ and $a_{j}$. Recall that $n_i = |L_i|$ is the number of distinct values the attribute $a_i$ assumes. The information about interconnection structure between two attributes $a_i$ and $a_j$ is summarized in a matrix of size $n_{i} \times n_{j}$ called {\em Mixing Matrix $M_{(a_{i},a_{j})}$}.  $M_{(a_{i},a_{j})}$ has a row (respectively column) corresponding to each distinct possible value of attribute $a_i$ (respectively $a_j)$. For $1\leq s \leq n_i$ and $1 \leq r \leq n_j$, the $(s,r)^{th}$ entry of  $M_{(a_{i},a_{j})}$ counts the number of edges (in the whole graph) connecting two nodes with attribute value $l_{s}$ of $a_{i}$ to nodes with attribute value $l_{r}$ of $a_{j}$. More precisely:
\begin{equation}\label{mixingMatrixDef}
    M_{(a_{i},a_{j})} (s,r) = \vert \{(u,v) \in E :  a_{i}(u) = l_{s} \text{ AND } a_{j}(v) = l_{r} \} \vert
\end{equation}
From construction of the mixing matrix it is clear that when values in $M_{(a_{i},a_{j})}$ are distributed more uniformly, then a given value of $a_i(v)$ does not reveal much information about $a_j(u)$ for some neighbor $u$ of $v$. However, if the distribution of numbers in $M_{(a_{i},a_{j})}$ is more skewed, then knowing the value of $a_i(v)$ one might be able to estimate the likelihood of a given value of $a_j(u)$ for some neighbor $u$ of $v$. 

The spread of values in $M_{(a_{i},a_{j})}$ are quantitatively measured by the {\em `divergence'} of $M_{(a_{i},a_{j})}$. The divergence $D$ of a real matrix $M$ with respect to a function $f$ (an input parameter) is defined as:
\begin{equation}\label{eq_correlation}
    D_{f} = \dfrac{ \sum_{i}^{} [f(e_{i\cdot}) - \sum_{j}^{} f(e_{ij})] + \sum_{j}^{} [f(e_{\cdot j}) - \sum_{i}^{} f(e_{ij})]}{\sum_{i}^{} f(e_{i\cdot}) + \sum_{j}^{} f(e_{\cdot j}) - 2 \sum_{i}^{}\sum_{j}^{} f (\frac{e_{\cdot j} e_{i\cdot}}{e_{\cdot \cdot}} )},
\end{equation}
where $f$ is a generative function, which can take values $f = x^{2}, x^{3}$, or $x \ log x$. The term $e_{i\cdot}$ is the sum of values in the $i^{th}$ row, $e_{\cdot j}$ is the sum of values in the $j^{th}$ column, $e_{\cdot\cdot}$ is the sum of all entries of the matrix $M$.  

The proclivity value (self/cross) between a pair of attributes (based on $M_{(a_{i},a_{j})}$) is inversely proportional to the divergence of $M_{(a_{i},a_{j})}$. A quantitative measure called ``proclivity index for attributed networks" (\textsc{prone}) is proposed in \cite{Prone}.
\textsc{prone} value between two attributes $a_i$ and $a_j$ is defined as:
\begin{equation}\label{final_prone_value}
 \text{\textsc{prone}}_{(a_{i},a_{j})} := \rho_{(a_{i},a_{j})} := 1-D_{f}
\end{equation}

\subsection{Construction of \textsc{n-fvr}}
Given a target node attribute $a_i$ and an integer $h$, ($0\leq h < |V|$), \textsc{n-fvr} constructs the feature vector $R_{a_i}^h(v)$ of a node $v\in V$ with respect to the attribute $a_i$. In the following we give a formal description of $R_{a_i}^h(v)$. For $h=0$, the network structure is not considered, and $R^0(v)$ is generated using only node $v$'s attributes. We refer to this approach as ``No-Network-Structure (\textsc{nns})" and define it as:

\begin{equation}\label{eq_zero_hop}
    R_{a_{i}}^{0}(v) = A(v)
\end{equation}

For $h>0$, we need to consider the attributes of nodes that are at distance at most $h$ from $v$. 
\subsubsection{Weight of $h$-hop neighborhood}
It is more likely that immediate neighbors reveal more information about $v$ than distant neighbors.
Therefore, more importance should be given to immediate neighbors ($h =1$) as compared to far away neighbors ($h >1$). 
Value of an attribute $a_{j}$ of the sets of node $N^{h}(v)$ ($h>0$) is a ``weighted" sum of $A_{j} (N^{h}(v))$ (weighted by $w_{i}$), for $1 \leq i \leq h$. Here $w_{i}$ is a network specific parameter that quantifies the diminishing influence of increasingly farther neighbors  on $v$. More precisely:
\begin{equation}
    A_{j}^{h}(v) = \sum_{i=1}^{h} w_{i} \times A_{j}(N^{i}(v)), 
\end{equation}
where $A_{j}(\cdot)$ is as defined in \eqref{single_attribute_equation_SAQ}. 
\subsubsection{Weight of attributes}
As discussed above, attributes have varying {\em correlation} with other attributes of their neighbor. This is reflected in the construction of $R_{a_i}^h(v)$ by giving attributes weights based on their proclivities. For $h > 0$, $R_{a_{i}}^{h}(v)$ is the weighted concatenation (weight is $\rho_{(a_{i}, a_{j})}$) of attributes values of $N^{h}(v)$. This is formally defined as: 

\begin{equation}\label{nfvr_final}
    R_{a_{i}}^{h} (v) =  A_{1}^{h}(v) \times \rho_{(a_{i},a_{1})} \oplus A_{2}^{h}(v) \times \rho_{(a_{i},a_{2})} \oplus \cdots \oplus A_{\vert A \vert}^{h}(v) \times \rho_{(a_{i},a_{\vert A \vert})},
\end{equation}
where, $\oplus$ represents the concatenation operation. We call the technique presented in Equation \eqref{nfvr_final} as \textsc{n-fvr}. Also we design another approach called ``\textsc{nns} and \textsc{n-fvr}" or ``\textsc{nn-fvr}", which is simply the concatenation of the vectors generated from Equation \eqref{eq_zero_hop} and Equation \eqref{nfvr_final}. The equation for \textsc{nn-fvr} is given as:

\begin{equation}
    R_{a_{i}}^{h} (v) = R_{a_{i}}^{0} (v) \oplus R_{a_{i}}^{h} (v)
\end{equation}

\begin{figure}[h!]
    \centering
    \includegraphics[scale=0.45]
    {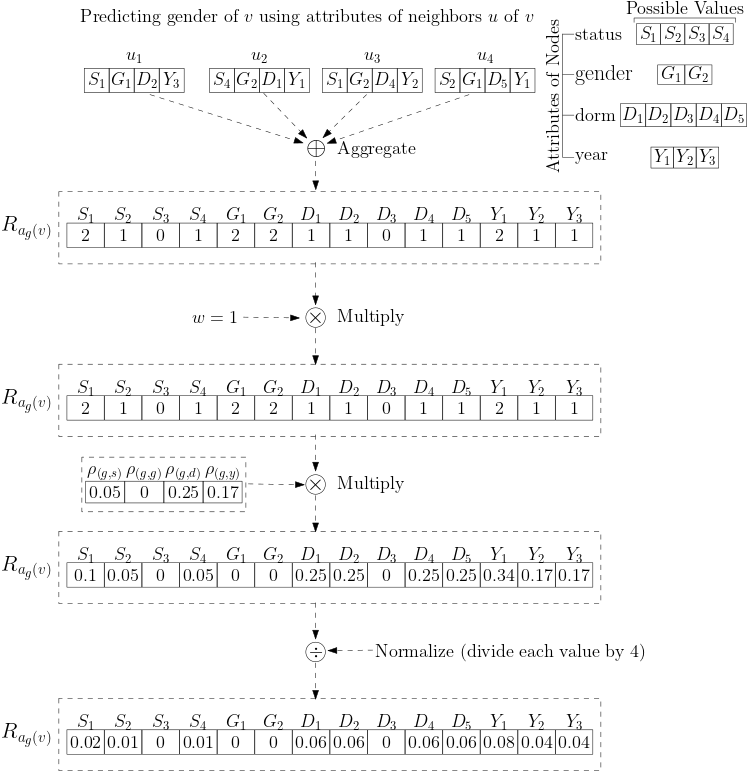}
    \caption{Graphical view for step by step working of Algorithm \ref{algo_proposed_approach} (\textsc{n-fvr}) for predicting gender $(g)$ of nodes in the {\em Caltech} dataset. Here we take $h$ = $1$ and $w = 1$. Note that all values in this example are dummy values for demonstration purposes. In ``Normalize $R_{v}$" step, values of each attribute are divided by $deg(v)$.}
    \label{algorithm_proposed_approach_graphic_view}
\end{figure}

Step by step procedure of our proposed approach is given in Algorithm \ref{algo_proposed_approach} and illustrated in Figure \ref{algorithm_proposed_approach_graphic_view}. In Algorithm \ref{algo_proposed_approach}, the For loops at line $3$ and $5$ are used to traverse the $h$ (hop length) and all attributes of nodes respectively.
Line $6$ aggregates the attributes values of neighbors (see Equation \eqref{h_hop_neighborhood_distance} and \eqref{single_attribute_equation_SAQ}) to generate a single feature vector with respect to an attribute and multiply attribute weight $\rho$ (see Equation \eqref{final_prone_value}) and hops weight $w$ with the feature vectors. Line $7$ concatenate the feature vectors of attributes and generate a separate feature vector for each value of $h$. These vectors are then aggregated in line $8$ to make a single feature vector for all levels of $h$. 
The feature vector is then normalized by dividing it with the degree of $v$ ($deg(v)$) in line $9$ to generate final feature vector representation for $v$ with respect to attribute $a_t$, for which missing value is being predicted.

\subsection{Runtime Analysis of \textsc{prone} and \textsc{n-fvr}}
For a graph $G=(V,E,A)$ with $|V|= n, |E| = m,$ and $|A|=t$, the mixing matrix (given in Equation \eqref{mixingMatrixDef}) can be populated in one linear scan over the edges in $G$. The divergence (given in Equation \eqref{eq_correlation}) of $M_{(a_i,a_j)}$ can be computed in $O(n_in_j)$ steps, where $n_i$ and $n_j$ are the number of levels of attributes $a_i$ and $a_j$, respectively. Thus \textsc{prone} (given in Equation \eqref{final_prone_value}) can be computed in $O(m + n_in_j)$ time. Note that $n_i \ll m$ (the number of edges in a graph is typically much larger than the number of levels of an attribute). Moreover, the prone value needs to be computed for each target attribute only once.

In case of \textsc{n-fvr}, feature vector for an attribute of node $v$ can be learned in $O(deg(v))$ (when $h=1$). For all nodes, this runtime is $O(m)$, where $m$ is the number of edges in the whole graph. 
For $h=2$, the runtime of a node $v$ is $O(deg(v) + \sum\limits_{u \in N(v)} deg(u))$. For $h=3$, the runtime of a node $v$ is $O(deg(v) + \sum\limits_{u \in N(v)} deg(u) + \sum\limits_{x \in N^{2}(v)} deg(x))$ and so on.

\begin{algorithm}[h!]
    \begin{algorithmic}[1]
        

        \For{$v \in V$}
        
        \State $R_{v}^{0} \gets A(v) $
        
        \For{$i \leftarrow 1:h $}\Comment{search depth}
        \State $h^{i} \gets []$
        \For{$j  \leftarrow 1:\vert A \vert $}
        \State $vec \gets w_{i} \times (\rho_{(a_{t}, a_{j})}) \times $ $(A_{j}(N^{i}(v)))$  \Comment{from Equation \eqref{h_hop_neighborhood_distance}, \eqref{single_attribute_equation_SAQ}, and \eqref{final_prone_value}}
        \State $y^{i} \gets$ \Call{Concat}{$y^{i},vec)$} 
        \EndFor\label{}
        \EndFor\label{}
        \State  y $\gets$ \Call{Aggregate}{$y^{1}$, $y^{2}$, $\ldots$ , $y^{h}$} 
        \State \Call{Normalize}{$y$} \Comment{divide by deg($v$)}
        \State $R1_{a_{t}}(v) \gets y$ \Comment{\textsc{n-fvr}}
        \State $R2_{a_{t}}(v) \gets $ \Call{Concat}{$R_{v}^{0},{y}$} \Comment{\textsc{nn-fvr}}
        
        \EndFor\label{} 
        \State \textbf{return} $R1, R2$
    \end{algorithmic}
    \caption{\textsc{n-fvr}(Graph $G = (V,E)$, $A(v), \forall \ v \in V$, hop length $h$, attributes weight $\rho$, hop weights $w$, attribute to be predicted $a_{t}$)}
    \label{algo_proposed_approach}
\end{algorithm}

\section{Experimental Setup}\label{experiments}
In this section, we describe the experimental setup, details of the ten benchmark datasets, comparison algorithms, the classification models we employed, and evaluation metrics.

\subsection{Datasets Description}
We use eight datasets from a collection of ``Facebook100" datasets, which are friendship networks of US universities \cite{traud2012social}. These datasets include ``Caltech", ``Rice", ``American", ``UChicago",  ``Mississippi", ``Temple", ``Haverford", and ``UNC". In each of these datasets, a node represents a user, and an edge between two nodes represents friendship. Each node has four associated categorical attributes, (i) status (faculty/student/etc.), (ii) gender, (iii) dormitory, and (iv) enrolled year. 

The next dataset is also a friendship network that is a subgraph of Pokec \cite{takac2012data}, a popular Slovak online social network. Each node in this graph has three associated attributes, (i) status (public/private), (ii) gender, and (iii) age. 

The last dataset is 4area, a bibliographic network extracted from \textsc{dblp} \cite{sun2009itopicmodel}. A node in this dataset represents an author and edges show co-author relationships. Each node has four attributes corresponding to four domains of computer science namely (i) database (\textsc{db}), (ii) data mining (\textsc{dm}), (iii) machine learning (\textsc{ml}), and (iv) information retrieval (\textsc{ir}). Attributes values are the fraction of research papers published by the authors in these areas. 
Note that attributes values in this dataset are continuous. For classification, we discretize attributes by binning into $5$ bins. However, for regression, we used the original continuous values. The statistics of datasets and training splits are presented in Table \ref{tbl:dataset_stats}. The train-test split is as in the literature, to ensure a fair comparison.

\begin{table}[h!]
	\footnotesize
	\centering
	\begin{tabular}{@{\extracolsep{4pt}}llccccc@{}}
		\toprule
		Dataset & Name & {No. of Nodes} & {No. of Edges} & {No. of Attributes} & {Train (\%)} \\
		\hline \hline
		 \multirow{8}{*}{Facebook100} & Caltech & $769$ & $33312$ & $4$ & 70 \\
		 & Haverford & $1446$ & $59589$ & $4$ & 1,5,9  \\
		 & Rice & $4088$ & $369657$ & $4$ & 70  \\
		 & American & $6387$ & $435325$ & $4$ & 70  \\
		 & UChicago & $6591$ & $208103$ & $4$ & 1,5,9  \\
		 & Mississippi & $10521$ & $610911$ & $4$ & 1,5,9  \\
		 & Temple & $13686$ & $360795$ & $4$ & 1,5,9  \\
		 & UNC & $18163$ & $766800$ & $4$ & 80  \\
		 Slovak Social Network & Pokec & $1000$ & $6303$ & $3$ & 70  \\
		 Bibliography Network & 4area & $26144$ & $217100$ & $4$ & 70  \\
		\bottomrule
	\end{tabular}
	\caption{Datasets statistics.}
	\label{tbl:dataset_stats}
\end{table}

\subsection{Comparison Algorithms}
We compare our approach with several methods that have been reported to have the best results on the corresponding datasets.
These methods are the following: 
\begin{enumerate}
\item \textbf{\textsc{\textsc{nns}}:} In this approach, network structure is not considered. The feature vector is generated using only the attributes of nodes (see Equation \eqref{eq_zero_hop}). 
\item \textbf{\textsc{wvrn} \cite{macskassy2003simple} (2003):} It is a weighted relational classifier that estimates attribute value $a_i$ of a node $v$ using the weighted mean of the same attribute of $v$'s neighbors. Since the graphs of the datasets are unweighted, we use similarity values to assign weights to the neighboring nodes. These similarity values are computed using Euclidean distance between the feature vector of node $v$ and its neighbors. 
\item \textbf{\textsc{lmmg} \cite{kim2012latent} (2012):} It is a node representation model, which is based upon the idea of Multiplicative Attribute Graph (MAG) Mode. In this approach, each node can belong to multiple groups, and the occurrence of each node feature is determined by a logistic model
based on the group memberships of the given node.
\item \textbf{DeepWalk \cite{perozzi2014deepwalk} (2014):} It is a representation model, which works for the unweighted graphs. The random walk method is used to translate graph structure into linear sequences. The skip-gram model with hierarchical softmax is used as the loss function. The code for DeepWalk is available online\footnote{\url{https://github.com/data-science-lab/data-science-lab.github.io/wiki/DeepWalk}}.
\item \textbf{\textsc{line} \cite{tang2015line} (2015):} It is a graph embedding approach, which preserves both local and global network structures and works for undirected, directed, and weighted graphs. It defines a loss function based on $1$-step and $2$-step relational information between nodes and combine them to get the final feature vector. The code for \textsc{line} is available online\footnote{\url{https://github.com/tangjianpku/LINE}}.
\item \textbf{GraRep \cite{cao2015grarep} (2015):} It is a node representation model for weighted graphs. This model incorporates both local and global structural information of the graph to learn the feature vector representations. The code for Grarep is available online\footnote{\url{https://github.com/benedekrozemberczki/GraRep}}.
\item \textbf{\textsc{slr} \cite{liao2016slr} (2016):} It is an integrative probabilistic model, which is used to capture the statistical correlations (homophily effect) among attributes. It uses the triangular motif representation of the network for improved scalability and predictive performance.
\item \textbf{node2vec \cite{grover2016node2vec} (2016):} It generalizes the DeepWalk method with the combination of BFS and DFS random walks. This method considers both network structure and graph homophily. The code for node2vec is available online\footnote{\url{https://snap.stanford.edu/node2vec/}}.
\item \textbf{\textsc{majority} \cite{duran2017learning} (2017):} The \textsc{majority} approach takes the most frequently occurring attribute values from the neighboring nodes in the training set and assigns that value to the attributes of nodes in the test set.
\item \textbf{\textsc{mne} \cite{yang2018multi} (2018):} This method captures multiple structures (facets) of the network by learning multiple embeddings simultaneously. It uses the Hilbert Schmidt Independence Criterion (HSIC) as a diversity constraint.

\end{enumerate}

\subsection{Prediction Models and Hyperparameters for \textsc{n-fvr} and \textsc{nn-fvr}}
To evaluate the goodness of the proposed approach, we use five standard classification/regression models. These include $k$-nearest neighbors (\textsc{knn}), na\"{\i}ve bayes (\textsc{nb}), decision tree (\textsc{dt}), support vector machine (\textsc{svm}), and linear regression (\textsc{lr}).

The hyperparameters of the feature map, $h$, and $w$ and those of each model are set using standard validation set approach.
Once these hyperparameters are tuned, unless otherwise mentioned, they remain same for all the experiments on all the datasets. The selected parameters are following:

\begin{itemize}
    \item Changing the value of $k$ for \textsc{knn} can greatly affect the predictive performance. We performed multiple experiments and empirically set $k=10$. This value remains the same across all datasets with the exception for Haverford, UChicago, Mississippi, and Temple datasets. For these datasets, the value of $k$ varies with respect to the attribute and size of training data. This was deemed necessary because of the different training splits (as opted in the baseline study~\cite{yang2018multi}). Values of $k$ for these datasets are given separately along with the results (see Table~\ref{tbl:mne_comparison}). Figure \ref{effect_of_k_caltech} shows the effect of changing $k$ on different attributes of Caltech dataset. Similar behavior is observed on other datasets.

\begin{figure}[h!]
	\centering
	\footnotesize
	\begin{tikzpicture}
	\begin{axis}[title={},
	compat=newest,
	xlabel style={text width=3.5cm, align=center},
	xlabel={{\small $k$}},
	ylabel={Accuracy (\%)}, ylabel shift={-3pt},xtick={},
	height=0.3\columnwidth, width=0.3\columnwidth, grid=major,
	ymin=0, ymax=100,
	]
	\node[text width=1cm] at (80,10) {status};
	\addplot+[
	mark size=1pt,
	smooth,
	error bars/.cd,
	y fixed,
	y dir=both,
	y explicit,
	] table [x={x}, y={y}, col sep=comma] {Data/k_effect/k_effect_for_attribute_status.csv};
	\end{axis}
	\end{tikzpicture}\hspace{-0.9cm}%
	\begin{tikzpicture}
	\begin{axis}[title={},
	compat=newest,
	xlabel style={text width=3.5cm, align=center},
	xlabel={{\small $k$}},
	yticklabels = {},
	height=0.3\columnwidth, width=0.3\columnwidth, grid=major,
	ymin=0, ymax=100,
	]
	\node[text width=1cm] at (80,10) {gender};
	\addplot+[
	mark size=1pt,
	smooth,
	error bars/.cd,
	y fixed,
	y dir=both,
	y explicit,
	] table [x={x}, y={y}, col sep=comma] {Data/k_effect/k_effect_for_attribute_gender.csv};
	\end{axis}
	\end{tikzpicture}\hspace{-0.9cm}%
	\begin{tikzpicture}
	\begin{axis}[title={},
	compat=newest,
	xlabel style={text width=3.5cm, align=center},
	xlabel={{\small $k$}},
	yticklabels = {},
	height=0.3\columnwidth, width=0.3\columnwidth, grid=major,
	ymin=0, ymax=100,
	]
	\node[text width=1cm] at (60,10) {dormitory};
	\addplot+[
	mark size=1pt,
	smooth,
	error bars/.cd,
	y fixed,
	y dir=both,
	y explicit,
	] table [x={x}, y={y}, col sep=comma] {Data/k_effect/k_effect_for_attribute_dorm.csv};
	\end{axis}
	\end{tikzpicture}\hspace{-0.9cm}%
	\begin{tikzpicture}
	\begin{axis}[title={},
	compat=newest,
	xlabel style={text width=3.5cm, align=center},
	xlabel={{\small $k$}},
	yticklabels = {},
	height=0.3\columnwidth, width=0.3\columnwidth, grid=major,
	ymin=0, ymax=100,
	]
	\node[text width=1cm] at (80,10) {year};
	\addplot+[
	mark size=1pt,
	smooth,
	error bars/.cd,
	y fixed,
	y dir=both,
	y explicit,
	] table [x={x}, y={y}, col sep=comma] {Data/k_effect/k_effect_for_attribute_year.csv};
	\end{axis}
	\end{tikzpicture}
	\caption{Effect of $k$ on accuracy using \textsc{knn} algorithm on different attributes of caltech dataset utilizing \textsc{n-fvr}.}
	\label{effect_of_k_caltech}
\end{figure}
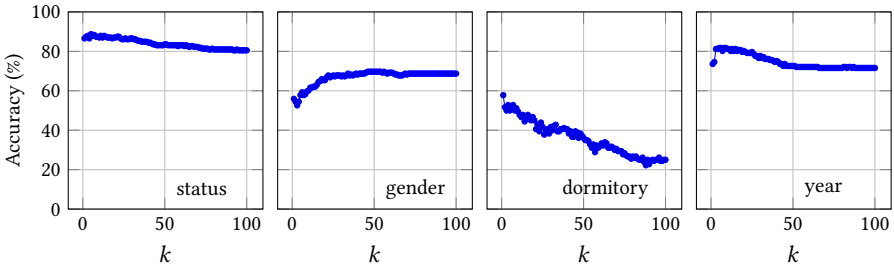

\item We only have one parameter for \textsc{nb} that is smoothing value, which is set to $0$.

\item The metric for root and attribute selection for \textsc{dt} is performed by utilizing ``gini index" value.

\item The kernel used for \textsc{svm} is ``linear" while classification type used is ``C-classification" with value of ``C" is taken as $1$.

\item We implement linear regression using ``\textsc{qr} matrix decomposition" instead of ``\textsc{svd} decomposition" due to its computational efficiency.

\item For each value of $h$, we use different weight $w$ so that equal importance is not given to immediate and far-away neighbors. We perform multiple experiments to select the value for $w$ from a range of $(0,1)$. We empirically decide the value for $w \in \{1, 0.5, 0.25\}$ and for $h \in \{1,2,3\}$.
\end{itemize}

\subsection{Evaluation Metrics and Implementation Details}
We use standard metrics from literature to evaluate the performance of our method. These metrics include accuracy, F1-measure, Mean Absolute Error (\textsc{mae}), Root Mean Squared Error (\textsc{rmse}), Mean Squared Error (\textsc{mse}), and $R^{2}$.

All experiments were carried out on a machine with an Intel(R) Core i$3$ CPU processor at $2.6$ GHz and $4$GB of DDR$3$ memory. Our code is implemented in \textit{R} (for feature vector generation and classification algorithms) and \textit{Weka} (for linear regression). 
The code is made publicly available for reproducibility and further experimentation  \footnote{\url{https://github.com/sarwanpasha/Attriubute-Prediction-Code}}.  

\section{Results and Comparisons} \label{Experimental_Evaluation}
This section presents the empirical evaluation of our model and its comparison with several baseline methods.
First, we present the heat map of the \textsc{prone} values for each pair of attributes of all datasets. 

\subsection{\textsc{prone} values ($\rho$)}
\begin{figure}[h!]
	\centering
	\begin{subfigure}{.175\textwidth}
		\centering
		\includegraphics[width=\linewidth]{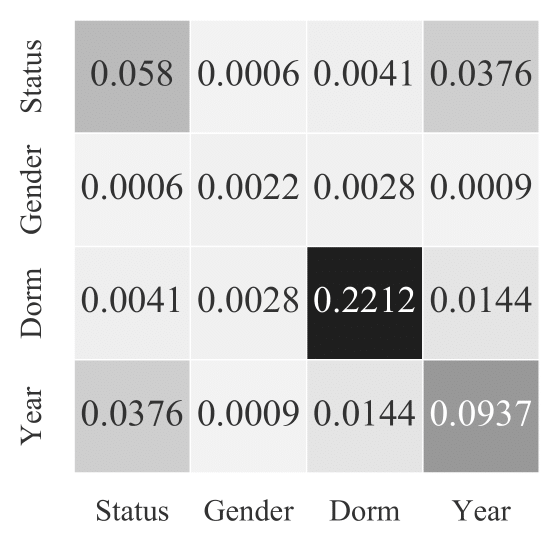}
		\caption{Caltech}
		\label{fig:sub_Caltech}
	\end{subfigure}%
	\begin{subfigure}{.175\textwidth}
		\centering
		\includegraphics[width=\linewidth]{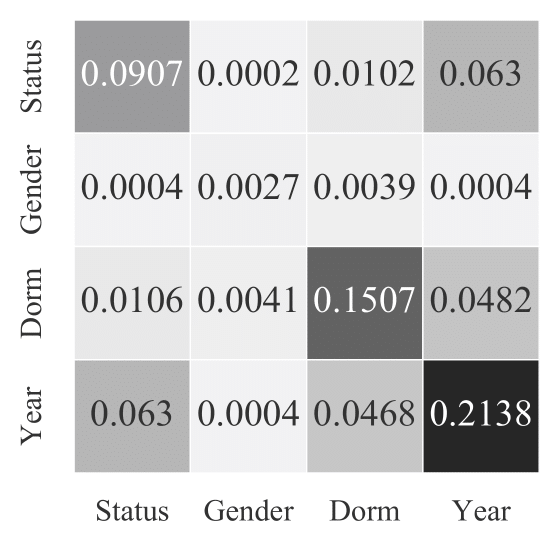}
		\caption{UNC}
		\label{fig:sub_unc}
	\end{subfigure}
	\begin{subfigure}{.175\textwidth}
		\centering
		\includegraphics[width=\linewidth]{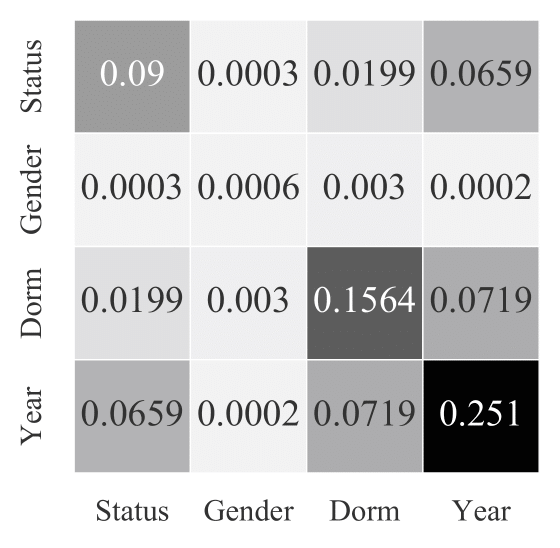}
		\caption{American}
		\label{fig:sub_American}
	\end{subfigure}
\begin{subfigure}{.175\textwidth}
	\centering
	\includegraphics[width=\linewidth]{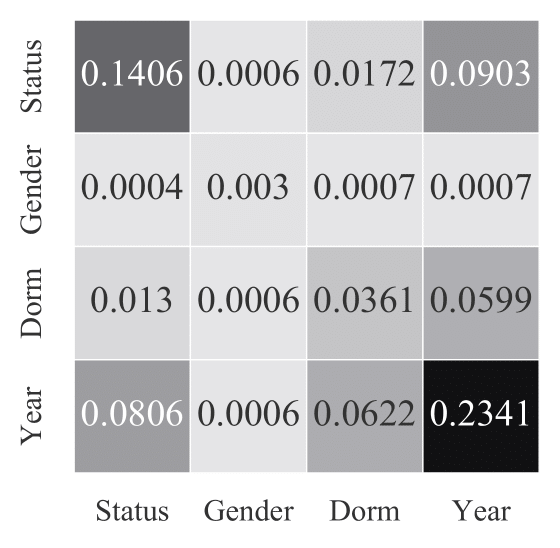}
	\caption{Haverford}
	\label{fig:sub_Haverford}
\end{subfigure}
	\begin{subfigure}{.225\textwidth}
		\centering
		\includegraphics[width=\linewidth]{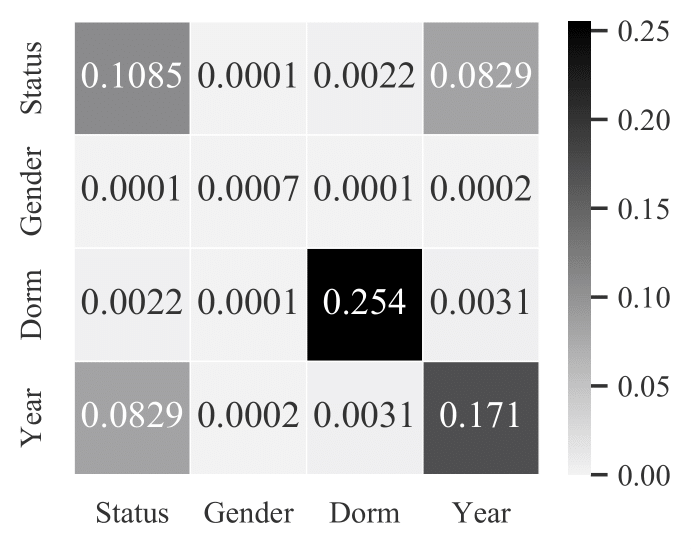}
		\caption{Rice}
		\label{fig:sub_rice}
	\end{subfigure}
	\begin{subfigure}{.175\textwidth}
		\centering
		\includegraphics[width=\linewidth]{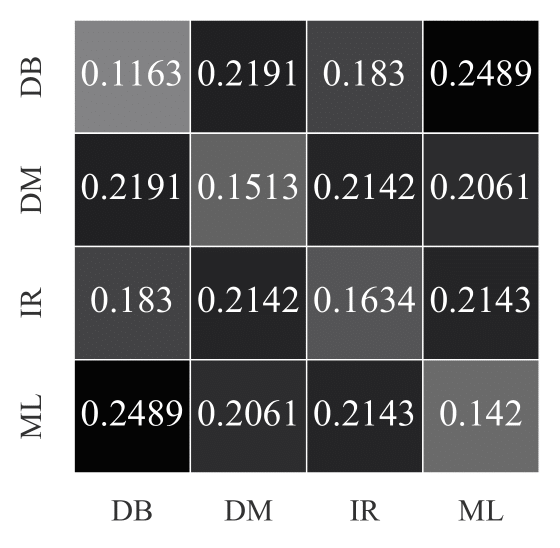}
		\caption{4area}
		\label{fig:sub_4area}
	\end{subfigure}
\begin{subfigure}{.175\textwidth}
	\centering
	\includegraphics[width=\linewidth]{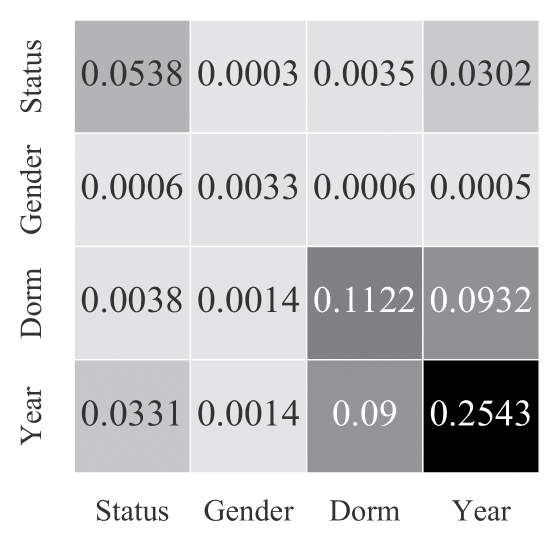}
	\caption{Temple}
	\label{fig:sub_Temple}
\end{subfigure}
\begin{subfigure}{.175\textwidth}
	\centering
	\includegraphics[width=\linewidth]{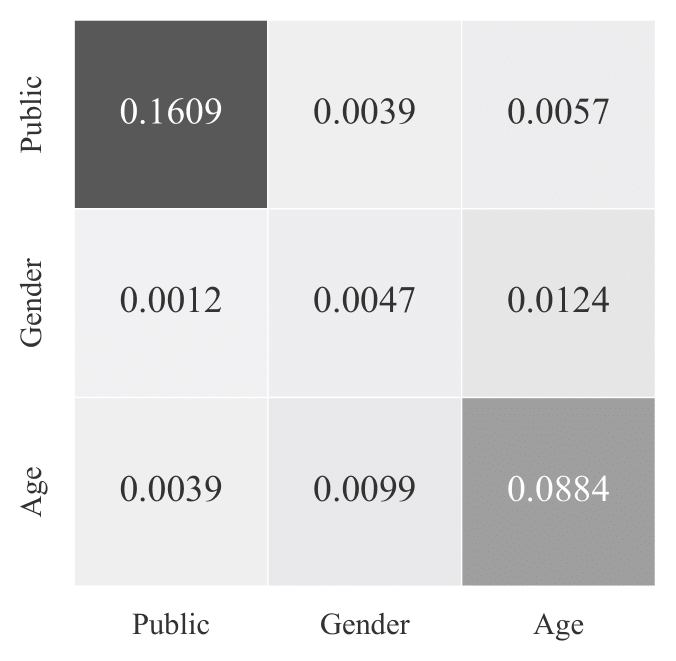}
	\caption{Pokec}
	\label{fig:sub_Pokec}
\end{subfigure}
\begin{subfigure}{.175\textwidth}
	\centering
	\includegraphics[width=\linewidth]{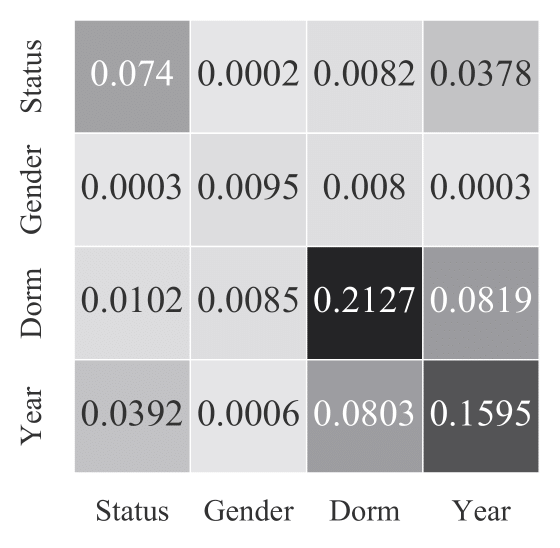}
	\caption{Mississippi}
	\label{fig:sub_Mississippi}
\end{subfigure}
\begin{subfigure}{.225\textwidth}
	\centering
	\includegraphics[width=\linewidth]{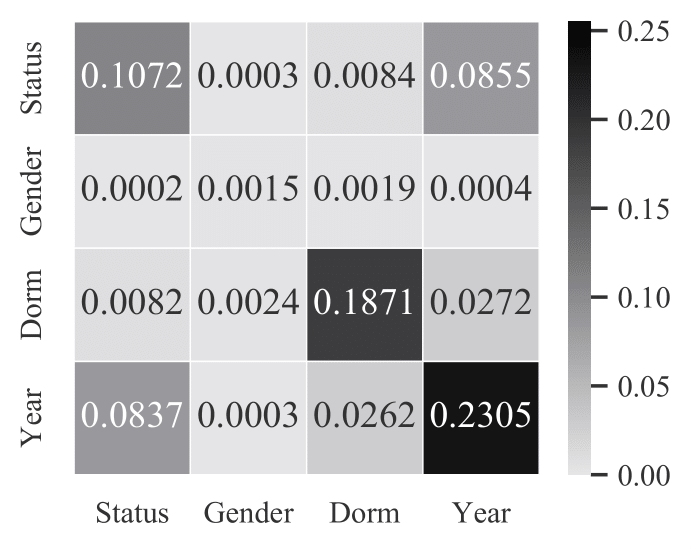}
	\caption{UChicago}
	\label{fig:sub_ucChicago}
\end{subfigure}
	\caption{\textsc{prone} values for the attributes of each dataset (which are used as attributes weights).}
	\label{fig:prone_Rice_Pokec_fourArea}
\end{figure}

We present the \textsc{prone} values $\rho$ between each pair of attributes for all datasets in Figure \ref{fig:prone_Rice_Pokec_fourArea}. The diagonal entries of these tables show the self-proclivity values for the corresponding attributes. The off-diagonal values show the cross-proclivity values between the corresponding pairs of attributes. These results are computed by taking value of generative function $f$ as $x \ log x$ in Equation \eqref{eq_correlation}. Identical trend is observed in case of $f = x^{2}$ and $f = x^{3}$.

As can be seen in Figure \ref{fig:prone_Rice_Pokec_fourArea}, dormitory attribute has the highest self proclivity in case of Caltech and Rice datasets (i.e., people belonging to the same dormitory tend to be friends). It essentially means that {\em given Alice and Bob are friends, if we know the dormitory value for Alice, then we can predict the dormitory value of Bob}.
However, gender shows very small self proclivity, therefore, the same cannot be said for it. This behavior is also observed for other facebook100 datasets. Interestingly, in the case of 4area dataset, many pairs of attributes pose high cross proclivity. This means that we can predict any attribute of 4area dataset using any other attribute of neighbors with high accuracy.

\subsection{Comparison with the Baseline Methods}
As mentioned earlier, we perform experiments by considering $3$ hop neighborhood. We first focus the discussion for $h = 1$.

The results for American and Rice datasets are presented in Table \ref{table:accuracy_comparison_American}. As can be seen, our methods significantly outperform the baselines (highest accuracy values are shown in boldface). However, the overall performance varies with respect to predicted attribute, classifier, and the \textsc{fvr} type used. 
The \textsc{dt} classifier on \textsc{nn-fvr} shows better performance  than other classifiers in case of ``status" and ``gender" attributes of both datasets. However, the highest accuracy is achieved with \textsc{svm} for ``dormitory" attribute. The \textsc{knn} algorithm with \textsc{n-fvr} show highest performance when predicting ``year" attribute of both datasets. However, these analyses do not help us to draw strong conclusions in favor of a particular setting of our experimentation. Hence, the results of one classifier on a particular attribute of a dataset cannot be generalized to all attributes. We also show the percentage improvement from \textsc{nns} to the best performing variation of the experiments utilizing \textsc{n-fvr} and \textsc{nn-fvr}. The highest gain in performance is observed for predicting ``dormitory" attribute using \textsc{nn-fvr} in case of American dataset ($76.62 \%$) and \textsc{n-fvr} in case of rice dataset ($83.67 \%$). 

\begin{table}[h!]
	\footnotesize
	\centering
	\begin{tabular}{p{2.4cm}p{0.58cm}p{0.58cm}p{1cm}p{0.58cm}cp{0.58cm}p{0.58cm}p{1cm}p{0.58cm}}
		
		\toprule
		\multirow{2}{*}{Method} &
		\multicolumn{4}{c}{American} & 
		\multicolumn{5}{c}{Rice}
		\\
		\cline{2-5} \cline{7-10}
		& {Status} & {Gender} & {Dormitory} & {Year} &    & {Status} & {Gender} & {Dormitory} & {Year} \\
		\hline \hline
		\textsc{wvrn}& 85.49 & 56.67 & 67.06 & 71.71  & & 86.12 & 54.80 & 84.46 & 74.72  \\
		\textsc{majority}& 85.43 & 56.84 & 67.11 & 70.99  & & 85.71 & 55 & 83.50 & 73.64\\
		\textsc{nns} & 79.59 &    59.70 &    16.93 &    38.96 & & 70.33 & 57.22 & 10.65 & 29.79 \\
		\hline
		\textsc{n-fvr \textsc{knn}} & 88.56 & 62.86 & {70.66} & \textbf{83.45}  & & 89.07 & 62.02 & \textbf{94.29} & \textbf{84.91} \\
		\textsc{n-fvr nb} & 80.16 &    62.86 &    36.79 &    48.63  & & 76.28 & 52.70 & 78.59 & 50.27 \\
		\textsc{n-fvr \textsc{dt}} & 88.51 &    62.86 &    54.23 &    81.55  & & 87.85 & 52.70 & 92.24 & 81.71\\
		\textsc{n-fvr \textsc{svm}} & 80.84 &    62.86 &    43.44 &    81.25  & & 84.59 & 52.70 & \textbf{94.29} & 77.33  \\
		\hline
		\textsc{nn-fvr \textsc{knn}} & 80.01 & 59.92 & 21.57 &  61.80  &  & 75.55 & 57.22 & 45.70 & 47.80  \\
		\textsc{nn-fvr nb} & 76.36  & 61.23 & 79.06 & 50.54  &  & 80.16 & 60.80 & 36.79 & 48.87  \\
		\textsc{nn-fvr \textsc{dt}} & \textbf{91.60} & \textbf{64.89} & 92.24  & 82.54  &  & \textbf{90.24} & \textbf{67.68} & 55.14 & {83.51} \\
		\textsc{nn-fvr \textsc{svm}} & 85 & 52.52 & \textbf{93.55} & 78.88  & & 81.41 & 62.86 & 43.85 & 81.43     \\
		\hline
		Improvement from \textsc{nns} to \textsc{n-fvr} (\%)  & \multirow{2}{*}{8.97} & \multirow{2}{*}{3.16} & \multirow{2}{*}{53.73} & \multirow{2}{*}{44.49}  & & \multirow{2}{*}{18.74}& \multirow{2}{*}{4.8} & \multirow{2}{*}{83.64} & \multirow{2}{*}{55.12}  \\
		Improvement from \textsc{nns} to \textsc{nn-fvr} (\%) & \multirow{2}{*}{10.65} & \multirow{2}{*}{5.19} & \multirow{2}{*}{76.62} & \multirow{2}{*}{43.58}   & & \multirow{2}{*}{19.91} & \multirow{2}{*}{10.46} & \multirow{2}{*}{44.49} & \multirow{2}{*}{53.72}\\
		\hline
	\end{tabular}
	\caption{Accuracy comparison of \textsc{n-fvr} and \textsc{nn-fvr} with \textsc{wvrn} \cite{macskassy2003simple} ($2003$), \textsc{majority} \cite{duran2017learning} ($2017$), and \textsc{nns} approaches on American and Rice dataset.}
	\label{table:accuracy_comparison_American}
\end{table}

The results for Pokec and 4area (after dividing attributes values into $5$ bins) datasets are shown in Table~\ref{table:accuracy_comparison_Pokec_dataset}. 
In regards to Pokec dataset, the \textsc{dt} classifier with \textsc{n-fvr} and \textsc{nn-fvr} shows highest performance for ``public" attribute. In case of ``gender" attribute, \textsc{knn} with \textsc{n-fvr} performs better. An interesting behavior is observed in case of ``age" attribute, where \textsc{wvrn} outperform both \textsc{n-fvr} and \textsc{nn-fvr} by a small margin.
Most interesting results are achieved on 4area dataset, on which the \textsc{svm} classifier with \textsc{nn-fvr} outperforms all other approaches for all the attributes. Overall, \textsc{nn-fvr} perform better than \textsc{nns} as evident from the mentioned performance gains. Note that for 4area dataset, every model and classifier achieve accuracy greater than $85 \%$. This is because of the fact that there is a high self/cross proclivity among attributes of 4area dataset (see Figure \ref{fig:prone_Rice_Pokec_fourArea}). It is observed that high self/cross proclivity enables all models to efficiently learn the patterns in the data, consequently leading to higher predictive performance. Therefore, we only observed minor performance gain for \textsc{nn-fvr} model (and negative gain for \textsc{n-fvr} model). However, since high proclivity is usually not observed in real-world scenarios (as can be seen in Figure \ref{fig:prone_Rice_Pokec_fourArea}), it can be concluded that the proposed model is more applicable in real-world settings.
\begin{table}[h!]
	\footnotesize
	\centering
	\begin{tabular}{p{2.5cm}cccccccc}
		
		\toprule
		\multirow{2}{*}{Method} &
		\multicolumn{3}{c}{Pokec} & 
		\multicolumn{5}{c}{4area}
		\\
		\cline{2-4} \cline{6-9} 
		& {Public} & {Gender} & {Age} &    & {\textsc{db}} & {\textsc{dm}} & {\textsc{ir}} & {\textsc{ml}} \\
		\hline \hline
		\textsc{wvrn}& 46.2 &  42.2 &  \textbf{25.6} &  & 90.40 & 88.94 & 88.97 & 89.95   \\
		\textsc{majority}& 49.1 & 40.5 & 25.2  &   & 90.17 & 88.84 & 88.57 & 89.68 \\
		\textsc{nns} & 52.33 & 61.33 & 16.74 &  & 97.60 & 97.50 & 97.30 & 97.90 \\
		\hline
		\textsc{n-fvr \textsc{knn}} & 87 & \textbf{66} & 23.78 &  & 92.83 & 92.26 & 92.01 & 92.40 \\
		\textsc{n-fvr nb} & 86.33 & 60.66 & 18.50 &  & 88.71 & 87.21 & 87.65 & 88.05 \\
		\textsc{n-fvr \textsc{dt}} & \textbf{87.66} & 61 & 21.58 &  & 92.45 & 92.59 & 91.48 & 92.93 \\
		\textsc{n-fvr \textsc{svm}} & 81 & 57 & 14.09 &  & 92.64 & 92.47 & 91.80 & 93.11 \\
		\hline
		\textsc{nn-fvr \textsc{knn}} & 80.66 & 64.33 & 25.11  &  & 96.30 & 95.52 & 95.46 & 96.18 \\
		\textsc{nn-fvr nb} & 86.33 & 60.66 & 18.94 &  & 90.04 &  89.31 & 89.02 & 90.88   \\
		\textsc{nn-fvr \textsc{dt}} & \textbf{87.66} & 65.33 & 18.50 &  & 95.80 & 95.49 & 95.06 & 95.69 \\
		\textsc{nn-fvr \textsc{svm}} & 82.66 & 57 & 16.74 &  & \textbf{97.61} & \textbf{97.59} & \textbf{97.46} & \textbf{97.92}   \\
		\hline
		Improvement from \textsc{nns} to \textsc{n-fvr} (\%) & \multirow{2}{*}{35.33} & \multirow{2}{*}{4.67} & \multirow{2}{*}{7.04} &  & \multirow{2}{*}{-4.77} & \multirow{2}{*}{-4.91} & \multirow{2}{*}{-5.29} & \multirow{2}{*}{-4.79} \\
		Improvement from \textsc{nns} to \textsc{nn-fvr} (\%) & \multirow{2}{*}{35.33} & \multirow{2}{*}{4} & \multirow{2}{*}{8.37} &  & \multirow{2}{*}{0.01} & \multirow{2}{*}{0.09} & \multirow{2}{*}{0.16} & \multirow{2}{*}{0.02}  \\
		\hline
	\end{tabular}
	\caption{Accuracy comparison of \textsc{n-fvr} and \textsc{nn-fvr} with \textsc{wvrn} \cite{macskassy2003simple} ($2003$), \textsc{majority} \cite{duran2017learning} ($2017$), and \textsc{nns} approaches on Pokec and 4area dataset.}
	\label{table:accuracy_comparison_Pokec_dataset}
\end{table}
\begin{table}[h!]
	\footnotesize
	\centering
	\begin{tabular}{p{2.5cm}ccccc}
		
		\toprule
		\multirow{2}{*}{Method} &
		\multicolumn{4}{c}{Caltech}
		\\
		\cline{2-5} 
		& {Status} & {Gender} & {Dormitory} & {Year} \\
		\hline \hline
		\textsc{wvrn}& 80.88 & 60.46 & 74.25 & 67.23  \\
		\textsc{lmmg}& 77 & \_ &  \_ &  \_   \\
		\textsc{majority}& 79.97 & 59.03 & 72.82 & 65.14  \\
		\textsc{nns} & 69.69 & 65.40 & 18.33 & 31.97  \\
		\hline
		\textsc{n-fvr \textsc{knn}} & 87.01 & 62.08 & \textbf{88.88} &  \textbf{80.20}  \\
		\textsc{n-fvr nb} & 75.75 & 64.45 & 72.22 &  51.26  \\
		\textsc{n-fvr \textsc{dt}} & 86.58 & 64.45 & 82.22 &  74.61  \\
		\textsc{n-fvr \textsc{svm}} & 69.69 & \textbf{68.72} & 72.77 &  28.42  \\
		\hline
		\textsc{nn-fvr \textsc{knn}} & 70.99 & 65.40 & 28.33 & 36.04  \\
		\textsc{nn-fvr nb} & 75.75 & 64.45 & 72.22 & 51.77  \\
		\textsc{nn-fvr \textsc{dt}} & \textbf{92.20} & 64.45 & 82.22 & 78.17  \\
		\textsc{nn-fvr \textsc{svm}} & 71.42 & \textbf{68.72} & 54.44 & 38.57  \\
		\hline
		
		Improvement from \textsc{nns} to \textsc{n-fvr} (\%) & \multirow{2}{*}{17.32} & \multirow{2}{*}{3.32} & \multirow{2}{*}{70.55} & \multirow{2}{*}{48.23} \\
		Improvement from \textsc{nns} to \textsc{nn-fvr} (\%) & \multirow{2}{*}{22.51} & \multirow{2}{*}{3.32} & \multirow{2}{*}{63.89} & \multirow{2}{*}{46.20} \\
		\hline
	\end{tabular}
	\caption{Accuracy comparison of \textsc{n-fvr} and \textsc{nn-fvr} with \textsc{wvrn} \cite{macskassy2003simple} ($2003$),\textsc{lmmg}  \cite{kim2012latent} ($2016$), \textsc{majority} \cite{duran2017learning} ($2017$), and \textsc{nns} on Caltech dataset.}
	\label{table:accuracy_comparison_fb100_Dataset_Caltech}
\end{table}

In Table \ref{table:accuracy_comparison_fb100_Dataset_Caltech}, we show the results for Caltech dataset. It is evident from the results that our method significantly outperforms existing approaches. The \textsc{dt} with \textsc{nn-fvr} significantly performs better than other approaches for prediction of status attribute. In the case of gender attribute, \textsc{svm} shows equal performance for both \textsc{n-fvr} and \textsc{nn-fvr} based approaches. While for other two attributes, \textsc{knn} classifier on \textsc{n-fvr} yields maximum accuracy. As compared to baseline models, the proposed approach yields noticeable performance improvements. Note that we took the accuracy results of the \textsc{lmmg} approach from the original study \cite{kim2012latent}, where the authors has reported the results for one attribute only (other attributes are left empty in Table \ref{table:accuracy_comparison_fb100_Dataset_Caltech}).

Results in Table \ref{table:accuracy_comparison_unc_chapel_hill} show the F1-score of our method and other baseline approaches on UNC dataset. We report $F1$-score to make fair comparison with the results given in \cite{liao2016slr} on the respective dataset.
We can see in the results that our method significantly outperforms all baseline approaches. The F1-score of the \textsc{slr} and \textsc{svd++} approaches were mentioned for only one attribute ``status" in the original study (other attributes are left empty in Table \ref{table:accuracy_comparison_unc_chapel_hill}).
As far as attribute specific performance is concerned, \textsc{dt} with \textsc{nn-fvr} shows highest F1-score while predicting ``status" and ``dormitory" attributes. Similarly, \textsc{svm} based on \textsc{n-fvr} yields maximum performance while predicting the ``gender" and ``year" attributes.
\begin{table}[h!]
	\footnotesize
	\centering
	\begin{tabular}{p{2.5cm}ccccc}
		
		\toprule
		\multirow{2}{*}{Method} &
		\multicolumn{4}{c}{UNC}
		\\
		\cline{2-5} 
		& {Status} & {Gender} & {Dormitory} & {Year} \\
		\hline \hline
		\textsc{wvrn}& 51.57 & 40.12 & 29.46 & 48.16   \\
		\textsc{slr-e} & 43 & \_ &  \_ &  \_ \\
		\textsc{slr-m} & 41 & \_ &  \_ &  \_ \\
		\textsc{svd++} & 57 & \_ &  \_ &  \_ \\
		\textsc{majority} & 49.87 & 39.22 & 26.17 & 45.40  \\
		\textsc{nns} & 40.20 & 57.77 & 7.53 & 33.48 \\
		\hline
		\textsc{n-fvr \textsc{knn}} & 69.14 & 54.19 & 24.30 & 58.58 \\
		\textsc{n-fvr nb} & 44.77 & 53.59 & 8.49 & 24.08 \\
		\textsc{n-fvr \textsc{dt}} & 67.74 & 57.96 & {35.07} & 68.74 \\
		\textsc{n-fvr \textsc{svm}} & {72.18} & \textbf{73.34} & 19.47 & \textbf{76.39} \\
		\hline 
		\textsc{nn-fvr \textsc{knn}} & 38.65 & 56.81 & 7.80 & 41.73  \\
		\textsc{nn-fvr nb} & 44.77 & 53.62 & 8.49 & 24.16  \\
		\textsc{nn-fvr \textsc{dt}} & \textbf{74.56} & {62.83} & \textbf{38.99} & {71.28}  \\
		\textsc{nn-fvr \textsc{svm}} & 73.78 & 43.07 & 9.69 & 73.70 \\ 
		\hline
		Improvement from \textsc{nns} to \textsc{n-fvr} (\%) & \multirow{2}{*}{31.98} & \multirow{2}{*}{15.57} & \multirow{2}{*}{27.54} & \multirow{2}{*}{42.91} \\
		Improvement from \textsc{nns} to \textsc{nn-fvr} (\%) & \multirow{2}{*}{34.36} & \multirow{2}{*}{5.06} & \multirow{2}{*}{31.46} & \multirow{2}{*}{40.22} \\
		\hline
	\end{tabular}
	\caption{F1-Score comparison of \textsc{n-fvr} and \textsc{nn-fvr} with \textsc{wvrn} \cite{macskassy2003simple} ($2003$), \textsc{slr-e} \cite{liao2016slr} ($2016$), \textsc{slr-m} \cite{liao2016slr} ($2016$), \textsc{svd++} \cite{liao2016slr} ($2016$), \textsc{majority} \cite{duran2017learning} ($2017$), and \textsc{nns} on UNC dataset.}
	\label{table:accuracy_comparison_unc_chapel_hill}
\end{table}

Recall that actual attributes values of 4area dataset are continuous. 
Now we present the results on continuous values and compare them with \textsc{nns}. Table \ref{table:Linear_regression_4_area_Dataset} show the \textsc{mae}, \textsc{rmse}, $R^{2}$, and \textsc{mse}  values for attributes of $4$area dataset. We apply linear regression on the feature vectors generated using \textsc{n-fvr} and \textsc{nn-fvr} approaches. Our method \textsc{nn-fvr} show comparable results with \textsc{nns}. We do not compare our methods with \textsc{wvrn} and \textsc{majority} because they do not work for continuous dataset.

\begin{table*}[h!]
	\footnotesize
	\centering
	\begin{tabular}{@{\extracolsep{2pt}}lp{0.25cm}p{0.25cm}p{0.25cm}p{0.25cm}p{0.25cm}p{0.25cm}p{0.25cm}p{0.25cm}p{0.25cm}p{0.25cm}p{0.25cm}p{0.25cm}p{0.25cm}p{0.25cm}p{0.25cm}p{0.25cm}p{0.2cm}@{}}
		
		\toprule
		\multirow{3}{*}{Method} &
		\multicolumn{16}{c}{4area} 
		\\
		\cline{2-17}
		\multirow{2}{*}{} &
		\multicolumn{4}{c}{\textsc{db}}  & 
		\multicolumn{4}{c}{\textsc{dm}}  & 
		\multicolumn{4}{c}{\textsc{ir}}  & 
		\multicolumn{4}{c}{\textsc{ml}}
		\\
		\cline{2-5} \cline{6-9} \cline{10-13} \cline{14-17}
		& {\textsc{mae}} & {\textsc{rmse}} & {$R^{2}$} & {\textsc{mse}} & {\textsc{mae}} & {\textsc{rmse}} & {$R^{2}$} & {\textsc{mse}}  & {\textsc{mae}} & {\textsc{rmse}} & {$R^{2}$} & {\textsc{mse}}  & {\textsc{mae}}& {\textsc{rmse}} & {$R^{2}$} & {\textsc{mse}}  \\
		\midrule \midrule
		\textsc{nns} & 0 & 0 & 1 & 0 & 0 & 0  & 1 & 0 & 0  & 0 & 1 & 0  & 0 & 0 & 1 & 0  \\
		\textsc{n-fvr} & 0.07 & 0.14 & 0.87 & 0.02 & 0.07 & 0.15  & 0.79 & 0.02 & 0.08  & 0.16 & 0.81 & 0.02  & 0.07 & 0.15 & 0.89 & 0.02  \\
		\textsc{nn-fvr} & 0 & 0 & 1 & 0 & 0 & 0  & 1 & 0 & 0  & 0 & 1 & 0  & 0 & 0 & 1 & 0 \\
		\bottomrule
	\end{tabular}
	\caption{\textsc{mae}, \textsc{rmse}, $R^{2}$, and \textsc{mse} error using linear regression on 4area dataset.}
	\label{table:Linear_regression_4_area_Dataset}
\end{table*}

To the best of our knowledge, Multi-Facet Network Embedding (\textsc{mne}) (2018) is the current best solution for our problem. \textsc{mne} has made comparisons with several generic techniques like DeepWalk \cite{perozzi2014deepwalk} (2014), \textsc{line} \cite{tang2015line} (2015), GraRep \cite{cao2015grarep} (2015), and node2vec \cite{grover2016node2vec} (2016). Yang et. al. in \cite{yang2018multi} already demonstrated (both theoretically and empirically) that \textsc{mne} outperformed those generic techniques in accuracy. Therefore, outperforming \textsc{mne} implies that our algorithm also outperforms the generic techniques. Thus, rather than separately comparing our model with each generic approach, we use the results reported in \cite{yang2018multi} to make the comparisons.
Results in Table~\ref{tbl:mne_comparison} show the comparison of our method with \textsc{mne}, DeepWalk, \textsc{line}, GraRep, and node2vec. Since the authors did not use ``status" attribute in their study in \cite{yang2018multi}, we also omit this particular attribute in our experiments.
To make a fair comparison, we selected the same percentage of the training set ($1 \%$, $5 \%$, and $9 \%$) as given in \cite{yang2018multi}. Note that contrary to previous results, we only use \textsc{knn} for this particular comparison (due to its higher performance).  Optimum values of $k$ for \textsc{knn} are selected empirically for each dataset, attribute, and training percentage. This was deemed necessary because of the unique setting of train split percentages. It is evident from the results given in Table~\ref{tbl:mne_comparison} that our approach (\textsc{n-fvr}) outperform all other baselines in most of the cases (up to $15.4 $\% improvement from \textsc{mne}) with exception in a few cases. 

\begin{table}[h!]
	\footnotesize
	\centering
	\begin{tabular}{@{\extracolsep{4pt}}p{1.5cm}p{0.17cm}p{0.17cm}p{0.17cm}p{0.17cm}p{0.17cm}p{0.17cm}p{0.17cm}p{0.17cm}p{0.17cm}p{0.17cm}p{0.17cm}p{0.17cm}p{0.17cm}p{0.17cm}p{0.17cm}p{0.17cm}p{0.17cm}p{0.17cm}c@{}} 
		
		\toprule
		\multirow{1}{*}{} &
		\multicolumn{9}{c}{UChicago}  & 
		\multicolumn{9}{c}{Temple} 
		\\
		\cline{2-10} \cline{11-19}
		\multirow{2}{*}{Techniques} &
		\multicolumn{3}{c}{Gender}  & 
		\multicolumn{3}{c}{Year} & 
		\multicolumn{3}{c}{Dormitory} & 
		\multicolumn{3}{c}{Gender}  & 
		\multicolumn{3}{c}{Year} & 
		\multicolumn{3}{c}{Dormitory} 
		\\
		\cline{2-4} \cline{5-7} \cline{8-10} \cline{11-13} \cline{14-16} \cline{17-19}
		& {$1 \%$} & {$5 \%$} & {$9 \%$} & {$1 \%$} & {$5 \%$} & {$9 \%$}  & {$1 \%$} & {$5 \%$} & {$9 \%$}	& {$1 \%$} & {$5 \%$} & {$9 \%$} & {$1 \%$} & {$5 \%$} & {$9 \%$}  & {$1 \%$} & {$5 \%$} & {$9 \%$}  &\\
		\midrule \midrule
		DeepWalk & 50.1 & 52.3 & 55.9 & 55.6 & 59.1 & 63.8 & 20.2 & 35.7 & 47.4 & 50.1 & 55.5 & 58.2 & 51.1 & 55.7 & 60.3 & 21.4 & 31.8 & 36.1  \\
		\textsc{line} & 52.1 & 54.1 & 56.9 & 61 & 61.9 & 65.2 & 21.1 & 43.5 & 50.1 & 52.9 & 57.9 & 58.5 & 56.3 & 66.9 & 69.6 & 25.4 & 32.7 & 38.2  \\
		GraRep & 47.7 & 48.5 & 50.1 & 50.5 & 55.3 & 59.9 & 18.6 & 30.3 & 40 & 45.6 & 49 & 55 & 50.3 & 57.2 & 65.1 & 21.7 & 29.6 & 31.5  \\
		node2vec & 51.3 & 53.5 & 55.2 & 60.2 & 61.2 & 64.1 & 22.1 & 39.8 & 49.7 & 51 & 54.8 & 57.9 & 52.8 & 55.3 & 64.2 & 20.2 & 29.8 & 38.1  \\
		\textsc{mne} & 54.5 & \textbf{57.7} & \textbf{59.7} & 58.1 & 65.9 & 67.7 & 24.8 & \textbf{48.2} & \textbf{54.4} & 55.9 & \textbf{61.4} & \textbf{62.9} & 61.5 & 69.9 & \textbf{72.7} & 30.1 & 36.1 & 41.9  \\
		\hline
		\textsc{n-fvr} & \textbf{55.8} & 56.6 & 56.7 & \textbf{70.7} & \textbf{74.3} & \textbf{75} & \textbf{25.8} & 46.1 & 54.1 & \textbf{57.3} & 57.5 & 58.3 & \textbf{69.4} & \textbf{70.3} & 70.6 & \textbf{37.8} & \textbf{45.4} & \textbf{48.4}  \\
		\textsc{nn-fvr} & 52.3 & 52.4 & 52.4 & 52.5 & 71.1 & 71 & 8 & 16.6 & 22.1 & 57 & 57.1 & 57.1 & 57.3 & 67.8 & 68 & 21.6 & 29.2 & 29.4 \\ \hline
		Value of $k$ & 12 & 97 & 3 & 4 & 4 & 17 & 1 & 1 & 1 & 89 & 45 & 24 & 26 & 18 & 25 & 6 & 9 & 10  \\ \hline
		Improvement from \textsc{mne} (\%) & \multirow{2}{*}{1.3} & \multirow{2}{*}{-1.1} & \multirow{2}{*}{-3} & \multirow{2}{*}{12.6} & \multirow{2}{*}{8.4} & \multirow{2}{*}{7.3} & \multirow{2}{*}{1} & \multirow{2}{*}{-2.1} & \multirow{2}{*}{-0.3} & \multirow{2}{*}{1.4} & \multirow{2}{*}{-3.9} & \multirow{2}{*}{-4.6} & \multirow{2}{*}{7.9} & \multirow{2}{*}{0.4} & \multirow{2}{*}{-2.1} & \multirow{2}{*}{7.7} & \multirow{2}{*}{9.3} & \multirow{2}{*}{6.5}  \\
		\bottomrule
	\end{tabular} 
	
	\begin{tabular}{@{\extracolsep{4pt}}p{1.5cm}p{0.17cm}p{0.17cm}p{0.17cm}p{0.17cm}p{0.17cm}p{0.17cm}p{0.17cm}p{0.17cm}p{0.17cm}p{0.17cm}p{0.17cm}p{0.17cm}p{0.17cm}p{0.17cm}p{0.17cm}p{0.17cm}p{0.17cm}p{0.17cm}c@{}}
		
		\toprule
		\multirow{1}{*}{} &
		\multicolumn{9}{c}{Haverford}  & 
		\multicolumn{9}{c}{Mississippi} 
		\\
		\cline{2-10} \cline{11-19}
		\multirow{2}{*}{Techniques} &
		\multicolumn{3}{c}{Gender}  & 
		\multicolumn{3}{c}{Year} & 
		\multicolumn{3}{c}{Dormitory} & 
		\multicolumn{3}{c}{Gender}  & 
		\multicolumn{3}{c}{Year} & 
		\multicolumn{3}{c}{Dormitory} 
		\\
		\cline{2-4} \cline{5-7} \cline{8-10} \cline{11-13} \cline{14-16} \cline{17-19}
		& {$1 \%$} & {$5 \%$} & {$9 \%$} & {$1 \%$} & {$5 \%$} & {$9 \%$}  & {$1 \%$} & {$5 \%$} & {$9 \%$}	& {$1 \%$} & {$5 \%$} & {$9 \%$} & {$1 \%$} & {$5 \%$} & {$9 \%$}  & {$1 \%$} & {$5 \%$} & {$9 \%$}  & \\
		\midrule \midrule
		DeepWalk & 50.6 & 53.5 & 57.3 & 61.4 & 76.7 & 81.1 & 29 & 37.4 & 43.9 & 53.1 & 60.4 & 60.9 & 46.5 & 55.3 & 61.6 & 32.5 & 44.1 & 48.3 \\
		\textsc{line} & 50.1 & 51.6 & 52.9 & 59.1 & 76.1 & 80.5 & 27.9 & 36.6 & 41.5 & 55.3 & 62.7 & 64.7 & 48.6 & 58.9 & 63.2 & 34.2 & 48.9 & 53.4 \\
		GraRep & 48.8 & 51.1 & 51.9 & 57.4 & 72.1 & 77.5 & 29 & 39.8 & 42.9 & 44.6 & 48 & 52.9 & 42.7 & 48.3 & 49.2 & 32.5 & 45.9 & 52.1 \\
		node2vec & 51.3 & 57.1 & 57.1 & 57.6 & 75.6 & 79.1 & 29.2 & 41.4 & 43.8 & 52.6 & 59.8 & 59.8 & 47.2 & 56.8 & 60.1 & 31.3 & 39.5 & 44.1  \\
		\textsc{mne} & 54.2 & 59.6 & 62.0 & 66.9 & 81.3 & \textbf{84.4} & 33 & \textbf{45.7} & \textbf{47.6} & 58.9 & 65.9 & 68 & 53.3 & 59.4 & 63.8 & 38.7 & 53.7 & 56.7  \\
		\hline
		\textsc{n-fvr} & \textbf{63.8} & \textbf{64.2} & \textbf{63.9} & \textbf{78.9} & \textbf{81.9} & 83.4 & \textbf{38.3} & 41.6 & 47.5 & \textbf{63.2} & \textbf{67.1} & \textbf{68.7} & \textbf{68.7} & \textbf{68.4} & \textbf{68.8} & \textbf{43.9} & \textbf{56.6} & \textbf{60.8}  \\
		\textsc{nn-fvr} & 54.3 & 55.9 & 57.3 & 34.3 & 54 & 63.9 & 37.1 & 38.3 & 39.4 & 55 & 58 & 60.4 & 56.3 & 67.9 & 67.7 & 15.2 & 27.9 & 30  \\ \hline
		Value of $k$ & 7 & 29 & 48 & 1 & 3 & 6 & 10 & 9 & 7 & 1 & 4 & 6 & 12 & 38 & 42 & 4 & 5 & 14  \\
		\hline
		Improvement from \textsc{mne} (\%) & \multirow{2}{*}{9.6} & \multirow{2}{*}{4.6} & \multirow{2}{*}{1.9} & \multirow{2}{*}{12} & \multirow{2}{*}{0.6} & \multirow{2}{*}{-1} & \multirow{2}{*}{5.3} & \multirow{2}{*}{-4.1} & \multirow{2}{*}{-0.1} & \multirow{2}{*}{4.3} & \multirow{2}{*}{1.2} & \multirow{2}{*}{0.7} & \multirow{2}{*}{15.4} & \multirow{2}{*}{9} & \multirow{2}{*}{5} & \multirow{2}{*}{5.2} & \multirow{2}{*}{2.9} & \multirow{2}{*}{4.1}  \\
		\bottomrule
	\end{tabular}
	\caption{Accuracy comparison of \textsc{n-fvr} and \textsc{nn-fvr}  using \textsc{knn} classifier with different baselines namely DeepWalk \cite{perozzi2014deepwalk} (2014), \textsc{line} \cite{tang2015line} (2015), GraRep \cite{cao2015grarep} (2015), node2vec \cite{grover2016node2vec} (2016), and \textsc{mne} \cite{yang2018multi} (2018). Accuracy is computed by taking $1$ \%, $5$ \%, and $9$ \% data as train set and rest of data as test set.}
	\label{tbl:mne_comparison}
\end{table}

\begin{figure}[h!]
	\centering
	\footnotesize
	\begin{tikzpicture}
	\begin{axis}[title={},
	compat=newest,
	xlabel style={text width=3.5cm, align=center},
	xlabel={{}},
	ylabel={Accuracy (\%)}, ylabel shift={-3pt},xtick={},
	height=0.3\columnwidth, width=0.3\columnwidth, grid=major,
	ymin=20, ymax=100,
	xtick={1,2,3},
	xticklabels = {},
	legend style={font=\tiny,draw=none,fill=none},
	legend entries={KNN,NB,DT,SVM},
	legend columns = -1,
	legend to name=commonlegend_mape_comparison
	]
	\node[text width=1cm] at (2.5,30) {status};
	\addplot+[
	mark size=2.5pt,
	smooth,
	error bars/.cd,
	y fixed,
	y dir=both,
	y explicit,
	] table [x={x}, y={Status_NFVR_KNN}, col sep=comma] {Data/hop_comparision_Caltech.csv};
	\addplot+[mark=triangle*,
	mark size=2.5pt,
	dashed,
	error bars/.cd,
	y fixed,
	y dir=both,
	y explicit
	] table [x={x}, y={Status_NFVR_NB}, col sep=comma] {Data/hop_comparision_Caltech.csv};
	\addplot+[mark=halfcircle*,
	mark size=2.5pt,
	dashed,
	error bars/.cd,
	y fixed,
	y dir=both,
	y explicit
	] table [x={x}, y={Status_NFVR_DT}, col sep=comma] {Data/hop_comparision_Caltech.csv};
	\addplot+[mark=10-pointed star,
	mark size=2.5pt,
	dashed,
	error bars/.cd,
	y fixed,
	y dir=both,
	y explicit
	] table [x={x}, y={Status_NFVR_SVM}, col sep=comma] {Data/hop_comparision_Caltech.csv};
	\end{axis}
	\end{tikzpicture}\hspace{-0.1cm}%
	\begin{tikzpicture}
	\begin{axis}[title={},
	compat=newest,
	xlabel style={text width=3.5cm, align=center},
	xlabel={{}},
	ylabel={},
	height=0.3\columnwidth, width=0.3\columnwidth, grid=major,
	ymin=20, ymax=100,
	xtick={1,2,3},
	xticklabels = {},
	ytick={20,40,60,80,100},
	yticklabels = {},
	legend style={font=\tiny,draw=none,fill=none},
	legend entries={},
	]
	\node[text width=1cm] at (2.5,30) {gender};
	\addplot+[
	mark size=2.5pt,
	smooth,
	error bars/.cd,
	y fixed,
	y dir=both,
	y explicit
	] table [x={x}, y={Gender_NFVR_KNN}, col sep=comma] {Data/hop_comparision_Caltech.csv};
	\addplot+[mark=triangle*,
	mark size=2.5pt,
	dashed,
	error bars/.cd,
	y fixed,
	y dir=both,
	y explicit
	] table [x={x}, y={Gender_NFVR_NB}, col sep=comma] {Data/hop_comparision_Caltech.csv};
	\addplot+[mark=halfcircle*,
	mark size=2.5pt,
	dashed,
	error bars/.cd,
	y fixed,
	y dir=both,
	y explicit
	] table [x={x}, y={Gender_NFVR_DT}, col sep=comma] {Data/hop_comparision_Caltech.csv};
	\addplot+[mark=10-pointed star,
	mark size=2.5pt,
	dashed,
	error bars/.cd,
	y fixed,
	y dir=both,
	y explicit
	] table [x={x}, y={Gender_NFVR_SVM}, col sep=comma] {Data/hop_comparision_Caltech.csv};
	\end{axis}
	\end{tikzpicture}\hspace{-0.1cm}%
	\begin{tikzpicture}
	\begin{axis}[title={},
	compat=newest,
	xlabel style={text width=3.5cm, align=center},
	xlabel={{}},
	ylabel={},xtick={},
	height=0.3\columnwidth, width=0.3\columnwidth, grid=major,
	ymin=20, ymax=100,
	xtick={1,2,3},
	xticklabels = {},
	ytick={20,40,60,80,100},
	yticklabels = {},
	legend style={font=\tiny,draw=none,fill=none},
	legend entries={},
	]
	\node[text width=1cm] at (1.7,30) {dormitory};
	\addplot+[
	mark size=2.5pt,
	smooth,
	error bars/.cd,
	y fixed,
	y dir=both,
	y explicit
	] table [x={x}, y={Dorm_NFVR_KNN}, col sep=comma] {Data/hop_comparision_Caltech.csv};
	\addplot+[mark=triangle*,
	mark size=2.5pt,
	dashed,
	error bars/.cd,
	y fixed,
	y dir=both,
	y explicit
	] table [x={x}, y={Dorm_NFVR_NB}, col sep=comma] {Data/hop_comparision_Caltech.csv};
	\addplot+[mark=halfcircle*,
	mark size=2.5pt,
	dashed,
	error bars/.cd,
	y fixed,
	y dir=both,
	y explicit
	] table [x={x}, y={Dorm_NFVR_DT}, col sep=comma] {Data/hop_comparision_Caltech.csv};
	\addplot+[mark=10-pointed star,
	mark size=2.5pt,
	dashed,
	error bars/.cd,
	y fixed,
	y dir=both,
	y explicit
	] table [x={x}, y={Dorm_NFVR_SVM}, col sep=comma] {Data/hop_comparision_Caltech.csv};
	\end{axis}
	\end{tikzpicture}\hspace{-0.1cm}%
	\begin{tikzpicture}
	\begin{axis}[title={},
	compat=newest,
	xlabel style={text width=3.5cm, align=center},
	xlabel={{}},
	ylabel={},xtick={},
	height=0.3\columnwidth, width=0.3\columnwidth, grid=major,
	ymin=20, ymax=100,
	xtick={1,2,3},
	xticklabels = {},
	ytick={20,40,60,80,100},
	yticklabels = {},
	legend style={font=\tiny,draw=none,fill=none},
	legend entries={},
	]
	\node[text width=1cm] at (2.5,30) {year};
	\addplot+[
	mark size=2.5pt,
	smooth,
	error bars/.cd,
	y fixed,
	y dir=both,
	y explicit
	] table [x={x}, y={Year_NFVR_KNN}, col sep=comma] {Data/hop_comparision_Caltech.csv};
	\addplot+[mark=triangle*,
	mark size=2.5pt,
	dashed,
	error bars/.cd,
	y fixed,
	y dir=both,
	y explicit
	] table [x={x}, y={Year_NFVR_NB}, col sep=comma] {Data/hop_comparision_Caltech.csv};
	\addplot+[mark=halfcircle*,
	mark size=2.5pt,
	dashed,
	error bars/.cd,
	y fixed,
	y dir=both,
	y explicit
	] table [x={x}, y={Year_NFVR_DT}, col sep=comma] {Data/hop_comparision_Caltech.csv};
	\addplot+[mark=10-pointed star,
	mark size=2.5pt,
	dashed,
	error bars/.cd,
	y fixed,
	y dir=both,
	y explicit
	] table [x={x}, y={Year_NFVR_SVM}, col sep=comma] {Data/hop_comparision_Caltech.csv};
	\end{axis}
	\end{tikzpicture}\hspace{-0.1cm}%
	\linebreak
	\begin{tikzpicture}
	\begin{axis}[title={},
	compat=newest,
	xlabel={{\small $h$}},
	ylabel={Accuracy (\%)}, ylabel shift={-3pt},xtick={},
	height=0.3\columnwidth, width=0.3\columnwidth, grid=major,
	ymin=15, ymax=100,
	xtick={1,2,3},
	ytick={20,40,60,80,100},
	legend style={font=\tiny,draw=none,fill=none},
	legend entries={KNN,NB,DT,SVM},
	legend columns = -1,
	legend to name=commonlegend_mape_comparison
	]
	\node[text width=1cm] at (2.5,30) {status};
	\addplot+[
	mark size=2.5pt,
	smooth,
	error bars/.cd,
	y fixed,
	y dir=both,
	y explicit
	] table [x={x}, y={Status_NNFVR_KNN}, col sep=comma] {Data/hop_comparision_Caltech.csv};
	\addplot+[mark=triangle*,
	mark size=2.5pt,
	dashed,
	error bars/.cd,
	y fixed,
	y dir=both,
	y explicit
	] table [x={x}, y={Status_NNFVR_NB}, col sep=comma] {Data/hop_comparision_Caltech.csv};
	\addplot+[mark=halfcircle*,
	mark size=2.5pt,
	dashed,
	error bars/.cd,
	y fixed,
	y dir=both,
	y explicit
	] table [x={x}, y={Status_NNFVR_DT}, col sep=comma] {Data/hop_comparision_Caltech.csv};
	\addplot+[mark=10-pointed star,
	mark size=2.5pt,
	dashed,
	error bars/.cd,
	y fixed,
	y dir=both,
	y explicit
	] table [x={x}, y={Status_NNFVR_SVM}, col sep=comma] {Data/hop_comparision_Caltech.csv};
	\end{axis}
	\end{tikzpicture}\hspace{-0.1cm}%
	\begin{tikzpicture}
	\begin{axis}[title={},
	compat=newest,
	xlabel={{\small $h$}},
	ylabel={},xtick={},
	height=0.3\columnwidth, width=0.3\columnwidth, grid=major,
	ymin=15, ymax=100,
	xtick={1,2,3},
	ytick={20,40,60,80,100},
	yticklabels = {},
	legend style={font=\tiny,draw=none,fill=none},
	legend entries={},
	]
	\node[text width=1cm] at (2.5,30) {gender};
	\addplot+[
	mark size=2.5pt,
	smooth,
	error bars/.cd,
	y fixed,
	y dir=both,
	y explicit
	] table [x={x}, y={Gender_NNFVR_KNN}, col sep=comma] {Data/hop_comparision_Caltech.csv};
	\addplot+[mark=triangle*,
	mark size=2.5pt,
	dashed,
	error bars/.cd,
	y fixed,
	y dir=both,
	y explicit
	] table [x={x}, y={Gender_NNFVR_NB}, col sep=comma] {Data/hop_comparision_Caltech.csv};
	\addplot+[mark=halfcircle*,
	mark size=2.5pt,
	dashed,
	error bars/.cd,
	y fixed,
	y dir=both,
	y explicit
	] table [x={x}, y={Gender_NNFVR_DT}, col sep=comma] {Data/hop_comparision_Caltech.csv};
	\addplot+[mark=10-pointed star,
	mark size=2.5pt,
	dashed,
	error bars/.cd,
	y fixed,
	y dir=both,
	y explicit
	] table [x={x}, y={Gender_NNFVR_SVM}, col sep=comma] {Data/hop_comparision_Caltech.csv};
	\end{axis}
	\end{tikzpicture}\hspace{-0.1cm}%
	\begin{tikzpicture}
	\begin{axis}[title={},
	compat=newest,
	xlabel={{\small $h$}},
	ylabel={},xtick={},
	height=0.3\columnwidth, width=0.3\columnwidth, grid=major,
	ymin=15, ymax=100,
	xtick={1,2,3},
	ytick={20,40,60,80,100},
	yticklabels = {},
	legend style={font=\tiny,draw=none,fill=none},
	legend entries={},
	]
	\node[text width=1cm] at (2,30) {dormitory};
	\addplot+[
	mark size=2.5pt, 
	smooth,
	error bars/.cd,
	y fixed,
	y dir=both,
	y explicit
	] table [x={x}, y={Dorm_NNFVR_KNN}, col sep=comma] {Data/hop_comparision_Caltech.csv};
	\addplot+[mark=triangle*,
	mark size=2.5pt,
	dashed,
	error bars/.cd,
	y fixed,
	y dir=both,
	y explicit
	] table [x={x}, y={Dorm_NNFVR_NB}, col sep=comma] {Data/hop_comparision_Caltech.csv};
	\addplot+[mark=halfcircle*,
	mark size=2.5pt,
	dashed,
	error bars/.cd,
	y fixed,
	y dir=both,
	y explicit
	] table [x={x}, y={Dorm_NNFVR_DT}, col sep=comma] {Data/hop_comparision_Caltech.csv};
	\addplot+[mark=10-pointed star,
	mark size=2.5pt,
	dashed,
	error bars/.cd,
	y fixed,
	y dir=both,
	y explicit
	] table [x={x}, y={Dorm_NNFVR_SVM}, col sep=comma] {Data/hop_comparision_Caltech.csv};
	\end{axis}
	\end{tikzpicture}\hspace{-0.1cm}%
	\begin{tikzpicture}
	\begin{axis}[title={},
	compat=newest,
	xlabel={{\small $h$}},
	ylabel={},xtick={},
	height=0.3\columnwidth, width=0.3\columnwidth, grid=major,
	ymin=15, ymax=100,
	xtick={1,2,3},
	ytick={20,40,60,80,100},
	yticklabels = {},
	legend style={font=\tiny,draw=none,fill=none},
	legend entries={},
	]
	\node[text width=1cm] at (2.5,25) {year};
	\addplot+[
	mark size=2.5pt,
	smooth,
	error bars/.cd,
	y fixed,
	y dir=both,
	y explicit
	] table [x={x}, y={Year_NNFVR_KNN}, col sep=comma] {Data/hop_comparision_Caltech.csv};
	\addplot+[mark=triangle*,
	mark size=2.5pt,
	dashed,
	error bars/.cd,
	y fixed,
	y dir=both,
	y explicit
	] table [x={x}, y={Year_NNFVR_NB}, col sep=comma] {Data/hop_comparision_Caltech.csv};
	\addplot+[mark=halfcircle*,
	mark size=2.5pt,
	dashed,
	error bars/.cd,
	y fixed,
	y dir=both,
	y explicit
	] table [x={x}, y={Year_NNFVR_DT}, col sep=comma] {Data/hop_comparision_Caltech.csv};
	\addplot+[mark=10-pointed star,
	mark size=2.5pt,
	dashed,
	error bars/.cd,
	y fixed,
	y dir=both,
	y explicit
	] table [x={x}, y={Year_NNFVR_SVM}, col sep=comma] {Data/hop_comparision_Caltech.csv};
	\end{axis}
	\end{tikzpicture}
	\\	\ref{commonlegend_mape_comparison}
	\caption{Effect of $h$ on \textsc{n-fvr} (top) and \textsc{nn-fvr} (bottom) methods using different classifiers for different attributes of Caltech dataset. Figures are best seen in color.}
	\label{NFVR_Caltech_h_hop_comparision}
\end{figure}
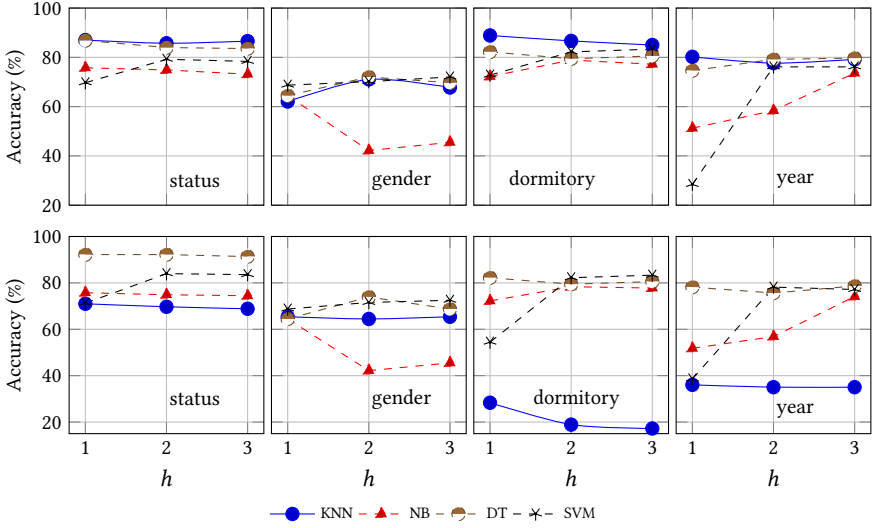
\subsection{Effect of $h$ on predictive performance}
Next, we investigate the effect of hop length $h$ on accuracy. Results in Figure~\ref{NFVR_Caltech_h_hop_comparision} show the accuracy with increasing value of $h$ for Caltech dataset. The results show a general trend of performance improvement (in most cases) when going from $h=1$ to $h=2$, with a few exceptions. However, going from $h = 2$ to $h = 3$, the performance gains are minimal.
Turning now to discussion about the specific classifiers, the performance of \textsc{svm} tend to increase for $h > 1$. 
On the other hand, on average, the performance of \textsc{nb} classifier tends to increase (except for gender attribute) with the increase in the value of $h$. 
The performance of \textsc{knn} does not have any noticeable effect in most cases.
These results also vary with datasets. Additional experiments reveal that going beyond $h = 1$ is dataset and attribute specific decision. No generic conclusion can be drawn with respect to these choices. For example, in Caltech dataset, it is evident from Figure \ref{NFVR_Caltech_h_hop_comparision} that there is no significant gain in the accuracy of status, dormitory, and year attributes in case of \textsc{knn} classifier for increasing value of $h$. However, we can observe slight improvement in accuracy of gender attribute in case of $h = 2$ (using \textsc{n-fvr}). 
From the results in Figure \ref{NFVR_Caltech_h_hop_comparision}, we can conclude that knowing the attributes values of neighbors of a node $v$ that are multiple hops away (i.e., $h>1$) provides no (or minimal) additional information for predicting the attributes of node $v$ (immediate neighbors have more influence on $v$). However, this conclusion does not hold for \textsc{svm} classifier, where we record significant performance boost for $h=2$. We observe similar behavior for other datasets as well.
Results of other datasets for increasing value of $h$ are shown in appendix~\ref{sec:appendix}.

\subsection{Effect of Self and Cross Proclivity}
Experimental results reveal that higher correlation amongst attributes results in higher predictive performance of the classifiers. The most noticeable effect of this phenomenon can be observed in case of 4area dataset, which has higher self/cross proclivity (see Figure \ref{fig:prone_Rice_Pokec_fourArea}). On this dataset all classifiers achieve accuracy greater than $85\%$ (see Table~\ref{table:accuracy_comparison_Pokec_dataset}). 
We observed that higher self or cross proclivity alone does not yield increased accuracy.  Effect of combining both proclivity measures can be observed for the datasets with low self/cross proclivity values among attributes. In this case, lower performance gains are achieved by the baseline methods including \textsc{nns}. However, since the proposed method is taking both self and cross proclivity into account, higher accuracy.
This accuracy vs. correlation behavior leads us to conclude that in a given dataset, if attributes show a high correlation amongst themselves (high self/cross proclivity), most predictive methods are very likely to predict attributes with higher accuracy and vice versa.


\subsection{Limitations}
Analysis of results reveals that when there is a high proclivity (self/cross) among the attributes, the proposed method does not perform significantly better than \textsc{nns}. This is evident from the results of 4area dataset (see Table \ref{table:accuracy_comparison_Pokec_dataset}). Even in this scenario, our method is competitive in majority of cases, but it does not significantly outperform existing methods. However, in majority of the real-world scenarios, high proclivity is not prevalent (see Figure \ref{fig:prone_Rice_Pokec_fourArea}). Secondly, we observe that as the number of unique values in attributes increases, the accuracy of underlying classifiers tends to decrease. This behavior is observed for all methods. 


\section{Conclusion and Future Work}\label{conclusion}
We propose a method to generate feature vectors for the nodes based on other attributes values of that node and its neighbors. These feature vectors then input to standard machine learning algorithms to predict attributes. Our approach efficiently predicts attributes with high accuracy and outperforms existing methods. Through extensive experimentation on several benchmark datasets, we also show that our approach works for different types of datasets, highlighting the proposed approach's generalizability. 
One possible future direction is to combine statistical-based learning algorithms with the proposed approach to design an ensemble technique to construct richer feature vector representations for node attributes. 
Another possible extension is to use the proposed method to design feature vectors for the nodes or graphs in general that can then be used for node or graph classification. An interesting future research direction is to make the weight parameter ${\mathbf w}$ a model parameter and learn it from the data. Similarly, more complex models' influence of neighborhoods on nodes' attributes is another possible extension of this work.











\bibliographystyle{ACM-Reference-Format}
\bibliography{attribute_prediction_main}

\pagebreak

\appendix

\section{Appendix} \label{sec:appendix}
\begin{figure}[!h]
	\centering
	\footnotesize
	\begin{tikzpicture}
	\begin{axis}[title={},
	compat=newest,
	xlabel style={text width=3.5cm, align=center},
	ylabel={Accuracy (\%)}, ylabel shift={-3pt},xtick={},
	height=0.3\columnwidth, width=0.3\columnwidth, grid=major,
	ymin=40, ymax=100,
	xtick={1,2,3},
	xticklabels = {},
	legend style={font=\tiny,draw=none,fill=none},
	legend entries={KNN,NB,DT,SVM},
	legend columns = -1,
	legend to name=commonlegend_mape_comparison
	]
	\node[text width=1cm] at (2.5,50) {status};
	\addplot+[
	mark size=2.5pt,
	smooth,
	error bars/.cd,
	y fixed,
	y dir=both,
	y explicit,
	] table [x={x}, y={Status_NFVR_KNN}, col sep=comma] {Data/hop_comparision_rice31.csv};
	\addplot+[mark=triangle*,
	mark size=2.5pt,
	dashed,
	error bars/.cd,
	y fixed,
	y dir=both,
	y explicit
	] table [x={x}, y={Status_NFVR_NB}, col sep=comma] {Data/hop_comparision_rice31.csv};
	\addplot+[mark=halfcircle*,
	mark size=2.5pt,
	dashed,
	error bars/.cd,
	y fixed,
	y dir=both,
	y explicit
	] table [x={x}, y={Status_NFVR_DT}, col sep=comma] {Data/hop_comparision_rice31.csv};
	\addplot+[mark=10-pointed star,
	mark size=2.5pt,
	dashed,
	error bars/.cd,
	y fixed,
	y dir=both,
	y explicit
	] table [x={x}, y={Status_NFVR_SVM}, col sep=comma] {Data/hop_comparision_rice31.csv};
	\end{axis}
	\end{tikzpicture}\hspace{-0.1cm}%
	\begin{tikzpicture}
	\begin{axis}[title={},
	compat=newest,
	xlabel style={text width=3.5cm, align=center},
	ylabel={},xtick={},
	height=0.3\columnwidth, width=0.3\columnwidth, grid=major,
	ymin=40, ymax=100,
	xtick={1,2,3},
	xticklabels = {},
	ytick={40,60,80,100},
	yticklabels = {},
	legend style={font=\tiny,draw=none,fill=none},
	legend entries={},
	]
	\node[text width=1cm] at (2.5,45) {gender};
	\addplot+[
	mark size=2.5pt,
	smooth,
	error bars/.cd,
	y fixed,
	y dir=both,
	y explicit
	] table [x={x}, y={Gender_NFVR_KNN}, col sep=comma] {Data/hop_comparision_rice31.csv};
	\addplot+[mark=triangle*,
	mark size=2.5pt,
	dashed,
	error bars/.cd,
	y fixed,
	y dir=both,
	y explicit
	] table [x={x}, y={Gender_NFVR_NB}, col sep=comma] {Data/hop_comparision_rice31.csv};
	\addplot+[mark=halfcircle*,
	mark size=2.5pt,
	dashed,
	error bars/.cd,
	y fixed,
	y dir=both,
	y explicit
	] table [x={x}, y={Gender_NFVR_DT}, col sep=comma] {Data/hop_comparision_rice31.csv};
	\addplot+[mark=10-pointed star,
	mark size=2.5pt,
	dashed,
	error bars/.cd,
	y fixed,
	y dir=both,
	y explicit
	] table [x={x}, y={Gender_NFVR_SVM}, col sep=comma] {Data/hop_comparision_rice31.csv};
	\end{axis}
	\end{tikzpicture}\hspace{-0.1cm}%
	\begin{tikzpicture}
	\begin{axis}[title={},
	compat=newest,
	xlabel style={text width=3.5cm, align=center},
	ylabel={},xtick={},
	height=0.3\columnwidth, width=0.3\columnwidth, grid=major,
	ymin=40, ymax=100,
	xtick={1,2,3},
	xticklabels = {},
	ytick={40,60,80,100},
	yticklabels = {},
	legend style={font=\tiny,draw=none,fill=none},
	legend entries={},
	]
	\node[text width=1cm] at (2,50) {dormitory};
	\addplot+[
	mark size=2.5pt,
	smooth,
	error bars/.cd,
	y fixed,
	y dir=both,
	y explicit
	] table [x={x}, y={Dorm_NFVR_KNN}, col sep=comma] {Data/hop_comparision_rice31.csv};
	\addplot+[mark=triangle*,
	mark size=2.5pt,
	dashed,
	error bars/.cd,
	y fixed,
	y dir=both,
	y explicit
	] table [x={x}, y={Dorm_NFVR_NB}, col sep=comma] {Data/hop_comparision_rice31.csv};
	\addplot+[mark=halfcircle*,
	mark size=2.5pt,
	dashed,
	error bars/.cd,
	y fixed,
	y dir=both,
	y explicit
	] table [x={x}, y={Dorm_NFVR_DT}, col sep=comma] {Data/hop_comparision_rice31.csv};
	\addplot+[mark=10-pointed star,
	mark size=2.5pt,
	dashed,
	error bars/.cd,
	y fixed,
	y dir=both,
	y explicit
	] table [x={x}, y={Dorm_NFVR_SVM}, col sep=comma] {Data/hop_comparision_rice31.csv};
	\end{axis}
	\end{tikzpicture}\hspace{-0.1cm}%
	\begin{tikzpicture}
	\begin{axis}[title={},
	compat=newest,
	xlabel style={text width=3.5cm, align=center},
	ylabel={},xtick={},
	height=0.3\columnwidth, width=0.3\columnwidth, grid=major,
	ymin=40, ymax=100,
	xtick={1,2,3},
	xticklabels = {},
	ytick={40,60,80,100},
	yticklabels = {},
	legend style={font=\tiny,draw=none,fill=none},
	legend entries={},
	]
	\node[text width=1cm] at (2.5,50) {year};
	\addplot+[
	mark size=2.5pt,
	smooth,
	error bars/.cd,
	y fixed,
	y dir=both,
	y explicit
	] table [x={x}, y={Year_NFVR_KNN}, col sep=comma] {Data/hop_comparision_rice31.csv};
	\addplot+[mark=triangle*,
	mark size=2.5pt,
	dashed,
	error bars/.cd,
	y fixed,
	y dir=both,
	y explicit
	] table [x={x}, y={Year_NFVR_NB}, col sep=comma] {Data/hop_comparision_rice31.csv};
	\addplot+[mark=halfcircle*,
	mark size=2.5pt,
	dashed,
	error bars/.cd,
	y fixed,
	y dir=both,
	y explicit
	] table [x={x}, y={Year_NFVR_DT}, col sep=comma] {Data/hop_comparision_rice31.csv};
	\addplot+[mark=10-pointed star,
	mark size=2.5pt,
	dashed,
	error bars/.cd,
	y fixed,
	y dir=both,
	y explicit
	] table [x={x}, y={Year_NFVR_SVM}, col sep=comma] {Data/hop_comparision_rice31.csv};
	\end{axis}
	\end{tikzpicture}\hspace{-0.1cm}%
	\linebreak
	\begin{tikzpicture}
	\begin{axis}[title={},
	compat=newest,
	xlabel={{\small $h$}},
	ylabel={Accuracy (\%)}, ylabel shift={-3pt},xtick={},
	height=0.3\columnwidth, width=0.3\columnwidth, grid=major,
	ymin=0, ymax=100,
	xtick={1,2,3},
	legend style={font=\tiny,draw=none,fill=none},
	legend entries={KNN,NB,DT,SVM},
	legend columns = -1,
	legend to name=commonlegend_mape_comparison
	]
	\node[text width=1cm] at (2.5,10) {status};
	\addplot+[
	mark size=2.5pt,
	smooth,
	error bars/.cd,
	y fixed,
	y dir=both,
	y explicit,
	] table [x={x}, y={Status_NNFVR_KNN}, col sep=comma] {Data/hop_comparision_rice31.csv};
	\addplot+[mark=triangle*,
	mark size=2.5pt,
	dashed,
	error bars/.cd,
	y fixed,
	y dir=both,
	y explicit
	] table [x={x}, y={Status_NNFVR_NB}, col sep=comma] {Data/hop_comparision_rice31.csv};
	\addplot+[mark=halfcircle*,
	mark size=2.5pt,
	dashed,
	error bars/.cd,
	y fixed,
	y dir=both,
	y explicit
	] table [x={x}, y={Status_NNFVR_DT}, col sep=comma] {Data/hop_comparision_rice31.csv};
	\addplot+[mark=10-pointed star,
	mark size=2.5pt,
	dashed,
	error bars/.cd,
	y fixed,
	y dir=both,
	y explicit
	] table [x={x}, y={Status_NNFVR_SVM}, col sep=comma] {Data/hop_comparision_rice31.csv};
	\end{axis}
	\end{tikzpicture}\hspace{-0.1cm}%
	\begin{tikzpicture}
	\begin{axis}[title={},
	compat=newest,
	xlabel={{\small $h$}},
	ylabel={},xtick={},
	height=0.3\columnwidth, width=0.3\columnwidth, grid=major,
	ymin=0, ymax=100,
	xtick={1,2,3},
	yticklabels = {},
	legend style={font=\tiny,draw=none,fill=none},
	legend entries={},
	]
	\node[text width=1cm] at (2.5,10) {gender};
	\addplot+[
	mark size=2.5pt,
	smooth,
	error bars/.cd,
	y fixed,
	y dir=both,
	y explicit
	] table [x={x}, y={Gender_NNFVR_KNN}, col sep=comma] {Data/hop_comparision_rice31.csv};
	\addplot+[mark=triangle*,
	mark size=2.5pt,
	dashed,
	error bars/.cd,
	y fixed,
	y dir=both,
	y explicit
	] table [x={x}, y={Gender_NNFVR_NB}, col sep=comma] {Data/hop_comparision_rice31.csv};
	\addplot+[mark=halfcircle*,
	mark size=2.5pt,
	dashed,
	error bars/.cd,
	y fixed,
	y dir=both,
	y explicit
	] table [x={x}, y={Gender_NNFVR_DT}, col sep=comma] {Data/hop_comparision_rice31.csv};
	\addplot+[mark=10-pointed star,
	mark size=2.5pt,
	dashed,
	error bars/.cd,
	y fixed,
	y dir=both,
	y explicit
	] table [x={x}, y={Gender_NNFVR_SVM}, col sep=comma] {Data/hop_comparision_rice31.csv};
	\end{axis}
	\end{tikzpicture}\hspace{-0.1cm}%
	\begin{tikzpicture}
	\begin{axis}[title={},
	compat=newest,
	xlabel={{\small $h$}},
	ylabel={},xtick={},
	height=0.3\columnwidth, width=0.3\columnwidth, grid=major,
	ymin=0, ymax=100,
	xtick={1,2,3},
	yticklabels = {},
	legend style={font=\tiny,draw=none,fill=none},
	legend entries={},
	]
	\node[text width=1cm] at (2.3,45) {dormitory};
	\addplot+[
	mark size=2.5pt,
	smooth,
	error bars/.cd,
	y fixed,
	y dir=both,
	y explicit
	] table [x={x}, y={Dorm_NNFVR_KNN}, col sep=comma] {Data/hop_comparision_rice31.csv};
	\addplot+[mark=triangle*,
	mark size=2.5pt,
	dashed,
	error bars/.cd,
	y fixed,
	y dir=both,
	y explicit
	] table [x={x}, y={Dorm_NNFVR_NB}, col sep=comma] {Data/hop_comparision_rice31.csv};
	\addplot+[mark=halfcircle*,
	mark size=2.5pt,
	dashed,
	error bars/.cd,
	y fixed,
	y dir=both,
	y explicit
	] table [x={x}, y={Dorm_NNFVR_DT}, col sep=comma] {Data/hop_comparision_rice31.csv};
	\addplot+[mark=10-pointed star,
	mark size=2.5pt,
	dashed,
	error bars/.cd,
	y fixed,
	y dir=both,
	y explicit
	] table [x={x}, y={Dorm_NNFVR_SVM}, col sep=comma] {Data/hop_comparision_rice31.csv};
	\end{axis}
	\end{tikzpicture}\hspace{-0.1cm}%
	\begin{tikzpicture}
	\begin{axis}[title={},
	compat=newest,
	xlabel={{\small $h$}},
	ylabel={},xtick={},
	height=0.3\columnwidth, width=0.3\columnwidth, grid=major,
	ymin=0, ymax=100,
	xtick={1,2,3},
	yticklabels = {},
	legend style={font=\tiny,draw=none,fill=none},
	legend entries={},
	]
	\node[text width=1cm] at (2.5,10) {year};
	\addplot+[
	mark size=2.5pt,
	smooth,
	error bars/.cd,
	y fixed,
	y dir=both,
	y explicit
	] table [x={x}, y={Year_NNFVR_KNN}, col sep=comma] {Data/hop_comparision_rice31.csv};
	\addplot+[mark=triangle*,
	mark size=2.5pt,
	dashed,
	error bars/.cd,
	y fixed,
	y dir=both,
	y explicit
	] table [x={x}, y={Year_NNFVR_NB}, col sep=comma] {Data/hop_comparision_rice31.csv};
	\addplot+[mark=halfcircle*,
	mark size=2.5pt,
	dashed,
	error bars/.cd,
	y fixed,
	y dir=both,
	y explicit
	] table [x={x}, y={Year_NNFVR_DT}, col sep=comma] {Data/hop_comparision_rice31.csv};
	\addplot+[mark=10-pointed star,
	mark size=2.5pt,
	dashed,
	error bars/.cd,
	y fixed,
	y dir=both,
	y explicit
	] table [x={x}, y={Year_NNFVR_SVM}, col sep=comma] {Data/hop_comparision_rice31.csv};
	\end{axis}
	\end{tikzpicture}\hspace{-0.1cm}%
	\\	\ref{commonlegend_mape_comparison}
	\caption{$h $-hop effect on \textsc{n-fvr} (top) and \textsc{nn-fvr} (bottom) method using different classifiers for different attributes of Rice dataset. Figures are best seen in color.}
	\label{NFVR_Rice31_h_hop_comparision}
\end{figure}
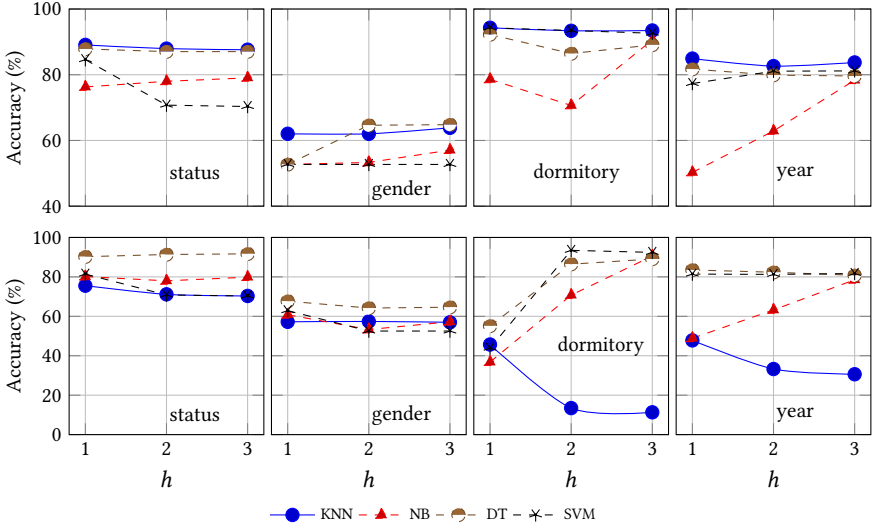

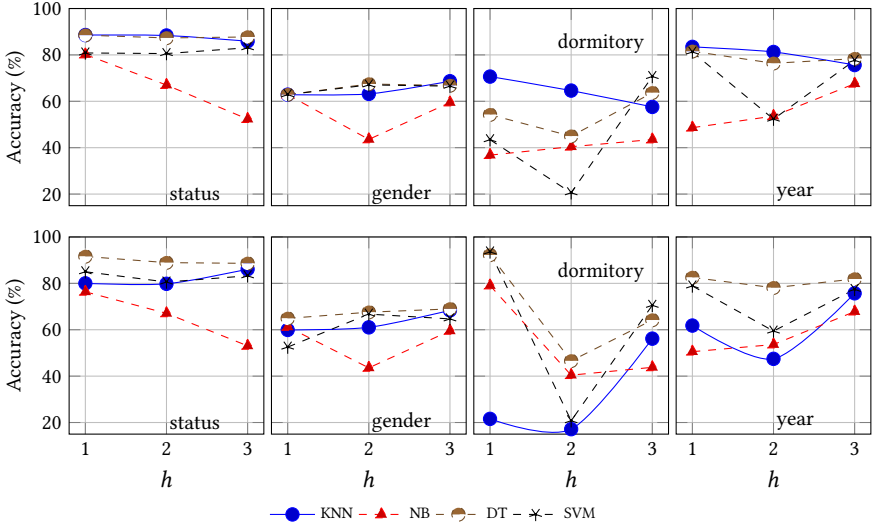
\begin{figure}[!h]
	\centering
	\footnotesize
	\begin{tikzpicture}
	\begin{axis}[title={},
	compat=newest,
	xlabel style={text width=3.5cm, align=center},
	ylabel={Accuracy (\%)}, ylabel shift={-3pt},xtick={},
	height=0.3\columnwidth, width=0.3\columnwidth, grid=major,
	ymin=15, ymax=100,
	xtick={1,2,3},
	xticklabels = {},
	legend style={font=\tiny,draw=none,fill=none},
	legend entries={KNN,NB,DT,SVM},
	legend columns = -1,
	legend to name=commonlegend_mape_comparison
	]
	\node[text width=1cm] at (2.5,20) {status};
	\addplot+[
	mark size=2.5pt,
	smooth,
	error bars/.cd,
	y fixed,
	y dir=both,
	y explicit,
	] table [x={x}, y={Status_NFVR_KNN}, col sep=comma] {Data/hop_comparision_american.csv};
	\addplot+[mark=triangle*,
	mark size=2.5pt,
	dashed,
	error bars/.cd,
	y fixed,
	y dir=both,
	y explicit
	] table [x={x}, y={Status_NFVR_NB}, col sep=comma] {Data/hop_comparision_american.csv};
	\addplot+[mark=halfcircle*,
	mark size=2.5pt,
	dashed,
	error bars/.cd,
	y fixed,
	y dir=both,
	y explicit
	] table [x={x}, y={Status_NFVR_DT}, col sep=comma] {Data/hop_comparision_american.csv};
	\addplot+[mark=10-pointed star,
	mark size=2.5pt,
	dashed,
	error bars/.cd,
	y fixed,
	y dir=both,
	y explicit
	] table [x={x}, y={Status_NFVR_SVM}, col sep=comma] {Data/hop_comparision_american.csv};
	\end{axis}
	\end{tikzpicture}\hspace{-0.1cm}%
	\begin{tikzpicture}
	\begin{axis}[title={},
	compat=newest,
	xlabel style={text width=3.5cm, align=center},
	ylabel={},xtick={},
	height=0.3\columnwidth, width=0.3\columnwidth, grid=major,
	ymin=15, ymax=100,
	xtick={1,2,3},
	xticklabels = {},
	ytick={20,40,60,80,100},
	yticklabels = {},
	legend style={font=\tiny,draw=none,fill=none},
	legend entries={},
	]
	\node[text width=1cm] at (2.5,20) {gender};
	\addplot+[
	mark size=2.5pt,
	smooth,
	error bars/.cd,
	y fixed,
	y dir=both,
	y explicit
	] table [x={x}, y={Gender_NFVR_KNN}, col sep=comma] {Data/hop_comparision_american.csv};
	\addplot+[mark=triangle*,
	mark size=2.5pt,
	dashed,
	error bars/.cd,
	y fixed,
	y dir=both,
	y explicit
	] table [x={x}, y={Gender_NFVR_NB}, col sep=comma] {Data/hop_comparision_american.csv};
	\addplot+[mark=halfcircle*,
	mark size=2.5pt,
	dashed,
	error bars/.cd,
	y fixed,
	y dir=both,
	y explicit
	] table [x={x}, y={Gender_NFVR_DT}, col sep=comma] {Data/hop_comparision_american.csv};
	\addplot+[mark=10-pointed star,
	mark size=2.5pt,
	dashed,
	error bars/.cd,
	y fixed,
	y dir=both,
	y explicit
	] table [x={x}, y={Gender_NFVR_SVM}, col sep=comma] {Data/hop_comparision_american.csv};
	\end{axis}
	\end{tikzpicture}\hspace{-0.1cm}%
	\begin{tikzpicture}
	\begin{axis}[title={},
	compat=newest,
	xlabel style={text width=3.5cm, align=center},
	ylabel={},xtick={},
	height=0.3\columnwidth, width=0.3\columnwidth, grid=major,
	ymin=15, ymax=100,
	xtick={1,2,3},
	xticklabels = {},
	ytick={20,40,60,80,100},
	yticklabels = {},
	legend style={font=\tiny,draw=none,fill=none},
	legend entries={},
	]
	\node[text width=1cm] at (2.3,85) {dormitory};
	\addplot+[
	mark size=2.5pt,
	smooth,
	error bars/.cd,
	y fixed,
	y dir=both,
	y explicit
	] table [x={x}, y={Dorm_NFVR_KNN}, col sep=comma] {Data/hop_comparision_american.csv};
	\addplot+[mark=triangle*,
	mark size=2.5pt,
	dashed,
	error bars/.cd,
	y fixed,
	y dir=both,
	y explicit
	] table [x={x}, y={Dorm_NFVR_NB}, col sep=comma] {Data/hop_comparision_american.csv};
	\addplot+[mark=halfcircle*,
	mark size=2.5pt,
	dashed,
	error bars/.cd,
	y fixed,
	y dir=both,
	y explicit
	] table [x={x}, y={Dorm_NFVR_DT}, col sep=comma] {Data/hop_comparision_american.csv};
	\addplot+[mark=10-pointed star,
	mark size=2.5pt,
	dashed,
	error bars/.cd,
	y fixed,
	y dir=both,
	y explicit
	] table [x={x}, y={Dorm_NFVR_SVM}, col sep=comma] {Data/hop_comparision_american.csv};
	\end{axis}
	\end{tikzpicture}\hspace{-0.1cm}%
	\begin{tikzpicture}
	\begin{axis}[title={},
	compat=newest,
	ylabel={},xtick={},
	height=0.3\columnwidth, width=0.3\columnwidth, grid=major,
	ymin=15, ymax=100,
	xtick={1,2,3},
	xticklabels = {},
	ytick={20,40,60,80,100},
	yticklabels = {},
	legend style={font=\tiny,draw=none,fill=none},
	legend entries={},
	]
	\node[text width=1cm] at (2.5,20) {year};
	\addplot+[
	mark size=2.5pt,
	smooth,
	error bars/.cd,
	y fixed,
	y dir=both,
	y explicit
	] table [x={x}, y={Year_NFVR_KNN}, col sep=comma] {Data/hop_comparision_american.csv};
	\addplot+[mark=triangle*,
	mark size=2.5pt,
	dashed,
	error bars/.cd,
	y fixed,
	y dir=both,
	y explicit
	] table [x={x}, y={Year_NFVR_NB}, col sep=comma] {Data/hop_comparision_american.csv};
	\addplot+[mark=halfcircle*,
	mark size=2.5pt,
	dashed,
	error bars/.cd,
	y fixed,
	y dir=both,
	y explicit
	] table [x={x}, y={Year_NFVR_DT}, col sep=comma] {Data/hop_comparision_american.csv};
	\addplot+[mark=10-pointed star,
	mark size=2.5pt,
	dashed,
	error bars/.cd,
	y fixed,
	y dir=both,
	y explicit
	] table [x={x}, y={Year_NFVR_SVM}, col sep=comma] {Data/hop_comparision_american.csv};
	\end{axis}
	\end{tikzpicture}\hspace{-0.1cm}%
	\linebreak
	\begin{tikzpicture}
	\begin{axis}[title={},
	compat=newest,
	xlabel={{\small $h$}},
	ylabel={Accuracy (\%)}, ylabel shift={-3pt},xtick={},
	height=0.3\columnwidth, width=0.3\columnwidth, grid=major,
	ymin=15, ymax=100,
	xtick={1,2,3},
	ytick={20,40,60,80,100},
	legend style={font=\tiny,draw=none,fill=none},
	legend entries={KNN,NB,DT,SVM},
	legend columns = -1,
	legend to name=commonlegend_mape_comparison
	]
	\node[text width=1cm] at (2.5,20) {status};
	\addplot+[
	mark size=2.5pt,
	smooth,
	error bars/.cd,
	y fixed,
	y dir=both,
	y explicit,
	] table [x={x}, y={Status_NNFVR_KNN}, col sep=comma] {Data/hop_comparision_american.csv};
	\addplot+[mark=triangle*,
	mark size=2.5pt,
	dashed,
	error bars/.cd,
	y fixed,
	y dir=both,
	y explicit
	] table [x={x}, y={Status_NNFVR_NB}, col sep=comma] {Data/hop_comparision_american.csv};
	\addplot+[mark=halfcircle*,
	mark size=2.5pt,
	dashed,
	error bars/.cd,
	y fixed,
	y dir=both,
	y explicit
	] table [x={x}, y={Status_NNFVR_DT}, col sep=comma] {Data/hop_comparision_american.csv};
	\addplot+[mark=10-pointed star,
	mark size=2.5pt,
	dashed,
	error bars/.cd,
	y fixed,
	y dir=both,
	y explicit
	] table [x={x}, y={Status_NNFVR_SVM}, col sep=comma] {Data/hop_comparision_american.csv};
	\end{axis}
	\end{tikzpicture}\hspace{-0.1cm}%
	\begin{tikzpicture}
	\begin{axis}[title={},
	compat=newest,
	xlabel={{\small $h$}},
	ylabel={},xtick={},
	height=0.3\columnwidth, width=0.3\columnwidth, grid=major,
	ymin=15, ymax=100,
	xtick={1,2,3},
	ytick={20,40,60,80,100},
	yticklabels = {},
	legend style={font=\tiny,draw=none,fill=none},
	legend entries={},
	]
	\node[text width=1cm] at (2.5,20) {gender};
	\addplot+[
	mark size=2.5pt,
	smooth,
	error bars/.cd,
	y fixed,
	y dir=both,
	y explicit
	] table [x={x}, y={Gender_NNFVR_KNN}, col sep=comma] {Data/hop_comparision_american.csv};
	\addplot+[mark=triangle*,
	mark size=2.5pt,
	dashed,
	error bars/.cd,
	y fixed,
	y dir=both,
	y explicit
	] table [x={x}, y={Gender_NNFVR_NB}, col sep=comma] {Data/hop_comparision_american.csv};
	\addplot+[mark=halfcircle*,
	mark size=2.5pt,
	dashed,
	error bars/.cd,
	y fixed,
	y dir=both,
	y explicit
	] table [x={x}, y={Gender_NNFVR_DT}, col sep=comma] {Data/hop_comparision_american.csv};
	\addplot+[mark=10-pointed star,
	mark size=2.5pt,
	dashed,
	error bars/.cd,
	y fixed,
	y dir=both,
	y explicit
	] table [x={x}, y={Gender_NNFVR_SVM}, col sep=comma] {Data/hop_comparision_american.csv};
	\end{axis}
	\end{tikzpicture}\hspace{-0.1cm}%
	\begin{tikzpicture}
	\begin{axis}[title={},
	compat=newest,
	xlabel={{\small $h$}},
	ylabel={},xtick={},
	height=0.3\columnwidth, width=0.3\columnwidth, grid=major,
	ymin=15, ymax=100,
	xtick={1,2,3},
	ytick={20,40,60,80,100},
	yticklabels = {},
	legend style={font=\tiny,draw=none,fill=none},
	legend entries={},
	]
	\node[text width=1cm] at (2.3,85) {dormitory};
	\addplot+[
	mark size=2.5pt,
	smooth,
	error bars/.cd,
	y fixed,
	y dir=both,
	y explicit
	] table [x={x}, y={Dorm_NNFVR_KNN}, col sep=comma] {Data/hop_comparision_american.csv};
	\addplot+[mark=triangle*,
	mark size=2.5pt,
	dashed,
	error bars/.cd,
	y fixed,
	y dir=both,
	y explicit
	] table [x={x}, y={Dorm_NNFVR_NB}, col sep=comma] {Data/hop_comparision_american.csv};
	\addplot+[mark=halfcircle*,
	mark size=2.5pt,
	dashed,
	error bars/.cd,
	y fixed,
	y dir=both,
	y explicit
	] table [x={x}, y={Dorm_NNFVR_DT}, col sep=comma] {Data/hop_comparision_american.csv};
	\addplot+[mark=10-pointed star,
	mark size=2.5pt,
	dashed,
	error bars/.cd,
	y fixed,
	y dir=both,
	y explicit
	] table [x={x}, y={Dorm_NNFVR_SVM}, col sep=comma] {Data/hop_comparision_american.csv};
	\end{axis}
	\end{tikzpicture}\hspace{-0.1cm}%
	\begin{tikzpicture}
	\begin{axis}[title={},
	compat=newest,
	xlabel={{\small $h$}},
	ylabel={},xtick={},
	height=0.3\columnwidth, width=0.3\columnwidth, grid=major,
	ymin=15, ymax=100,
	xtick={1,2,3},
	ytick={20,40,60,80,100},
	yticklabels = {},
	legend style={font=\tiny,draw=none,fill=none},
	legend entries={},
	]
	\node[text width=1cm] at (2.5,20) {year};
	\addplot+[
	mark size=2.5pt,
	smooth,
	error bars/.cd,
	y fixed,
	y dir=both,
	y explicit
	] table [x={x}, y={Year_NNFVR_KNN}, col sep=comma] {Data/hop_comparision_american.csv};
	\addplot+[mark=triangle*,
	mark size=2.5pt,
	dashed,
	error bars/.cd,
	y fixed,
	y dir=both,
	y explicit
	] table [x={x}, y={Year_NNFVR_NB}, col sep=comma] {Data/hop_comparision_american.csv};
	\addplot+[mark=halfcircle*,
	mark size=2.5pt,
	dashed,
	error bars/.cd,
	y fixed,
	y dir=both,
	y explicit
	] table [x={x}, y={Year_NNFVR_DT}, col sep=comma] {Data/hop_comparision_american.csv};
	\addplot+[mark=10-pointed star,
	mark size=2.5pt,
	dashed,
	error bars/.cd,
	y fixed,
	y dir=both,
	y explicit
	] table [x={x}, y={Year_NNFVR_SVM}, col sep=comma] {Data/hop_comparision_american.csv};
	\end{axis}
	\end{tikzpicture}\hspace{-0.1cm}%
	\\	\ref{commonlegend_mape_comparison}
	\caption{$h$-hop effect on \textsc{n-fvr} (top) and \textsc{nn-fvr} (bottom) method using different classifiers for different attributes of American dataset. Figures are best seen in color.}
	\label{NFVR_american_h_hop_comparision}
\end{figure}

The effect of using $h $-hop neighbors on all the prediction of all the attributes using multiple classifiers is presented here for the sake of completeness. This includes both \textsc{n-fvr} and \textsc{nn-fvr} based approaches. Figure~\ref{NFVR_Rice31_h_hop_comparision} shows the results for all three values of $h$-hop. In general, the performance of \textsc{knn} does not show any significant improvement when we increase the value of $h$ in case of \textsc{n-fvr}. However, in case of \textsc{nn-fvr}, \textsc{knn} shows decrease in performance if the value of $h$ is increased from $1$. The most interesting results are observed in case of \textsc{svm} and \textsc{nb} classifiers. Particularly, in case of dormitory attribute, \textsc{nb} and \textsc{svm} show a significant performance increment with $h$ $=2$. However, the time cost for each hop should also be considered. Therefore, this does not provide a conclusive evidence regarding the best choice for the value of $h$.

In regards to American dataset, the results for $h$-hop are presented in Figure~\ref{NFVR_american_h_hop_comparision}. In majority of the cases, we observe decrease in performance as we increase the value of $h$. This is true for both \textsc{n-fvr} and \textsc{nn-fvr} based experiments. Similar trends are observed in case of Pokec dataset, which are presented in Figure~\ref{NFVR_pokec_h_hop_comparision}.
\begin{figure}[t]
	\centering
	\footnotesize
	\begin{tikzpicture}
	\begin{axis}[title={},
	compat=newest,
	ylabel={Accuracy (\%)}, ylabel shift={-3pt},xtick={},
	height=0.3\columnwidth, width=0.3\columnwidth, grid=major,
	ymin=0, ymax=100,
	xtick={1,2,3},
	xticklabels = {},
	legend style={font=\tiny,draw=none,fill=none},
	legend entries={KNN,NB,DT,SVM},
	legend columns = -1,
	legend to name=commonlegend_mape_comparison
	]
	\node[text width=1cm] at (2.5,20) {public};
	\addplot+[
	mark size=2.5pt,
	smooth,
	error bars/.cd,
	y fixed,
	y dir=both,
	y explicit,
	] table [x={x}, y={Public_NFVR_KNN}, col sep=comma] {Data/hop_comparision_pokec.csv};
	\addplot+[mark=triangle*,
	mark size=2.5pt,
	dashed,
	error bars/.cd,
	y fixed,
	y dir=both,
	y explicit
	] table [x={x}, y={Public_NFVR_NB}, col sep=comma] {Data/hop_comparision_pokec.csv};
	\addplot+[mark=halfcircle*,
	mark size=2.5pt,
	dashed,
	error bars/.cd,
	y fixed,
	y dir=both,
	y explicit
	] table [x={x}, y={Public_NFVR_DT}, col sep=comma] {Data/hop_comparision_pokec.csv};
	\addplot+[mark=10-pointed star,
	mark size=2.5pt,
	dashed,
	error bars/.cd,
	y fixed,
	y dir=both,
	y explicit
	] table [x={x}, y={Public_NFVR_SVM}, col sep=comma] {Data/hop_comparision_pokec.csv};
	\end{axis}
	\end{tikzpicture}\hspace{-0.1cm}%
	\begin{tikzpicture}
	\begin{axis}[title={},
	compat=newest,
	ylabel={},
	height=0.3\columnwidth, width=0.3\columnwidth, grid=major,
	ymin=0, ymax=100,
	xtick={1,2,3},
	xticklabels = {},
	yticklabels = {},
	legend style={font=\tiny,draw=none,fill=none},
	legend entries={},
	]
	\node[text width=1cm] at (2.5,20) {gender};
	\addplot+[
	mark size=2.5pt,
	smooth,
	error bars/.cd,
	y fixed,
	y dir=both,
	y explicit
	] table [x={x}, y={Gender_NFVR_KNN}, col sep=comma] {Data/hop_comparision_pokec.csv};
	\addplot+[mark=triangle*,
	mark size=2.5pt,
	dashed,
	error bars/.cd,
	y fixed,
	y dir=both,
	y explicit
	] table [x={x}, y={Gender_NFVR_NB}, col sep=comma] {Data/hop_comparision_pokec.csv};
	\addplot+[mark=halfcircle*,
	mark size=2.5pt,
	dashed,
	error bars/.cd,
	y fixed,
	y dir=both,
	y explicit
	] table [x={x}, y={Gender_NFVR_DT}, col sep=comma] {Data/hop_comparision_pokec.csv};
	\addplot+[mark=10-pointed star,
	mark size=2.5pt,
	dashed,
	error bars/.cd,
	y fixed,
	y dir=both,
	y explicit
	] table [x={x}, y={Gender_NFVR_SVM}, col sep=comma] {Data/hop_comparision_pokec.csv};
	\end{axis}
	\end{tikzpicture}\hspace{-0.1cm}%
	\begin{tikzpicture}
	\begin{axis}[title={},
	compat=newest,
	ylabel={},
	height=0.3\columnwidth, width=0.3\columnwidth, grid=major,
	ymin=0, ymax=100,
	xtick={1,2,3},
	xticklabels = {},
	yticklabels = {},
	legend style={font=\tiny,draw=none,fill=none},
	legend entries={},
	]
	\node[text width=1cm] at (2.5,65) {age};
	\addplot+[
	mark size=2.5pt,
	smooth,
	error bars/.cd,
	y fixed,
	y dir=both,
	y explicit
	] table [x={x}, y={Age_NFVR_KNN}, col sep=comma] {Data/hop_comparision_pokec.csv};
	\addplot+[mark=triangle*,
	mark size=2.5pt,
	dashed,
	error bars/.cd,
	y fixed,
	y dir=both,
	y explicit
	] table [x={x}, y={Age_NFVR_NB}, col sep=comma] {Data/hop_comparision_pokec.csv};
	\addplot+[mark=halfcircle*,
	mark size=2.5pt,
	dashed,
	error bars/.cd,
	y fixed,
	y dir=both,
	y explicit
	] table [x={x}, y={Age_NFVR_DT}, col sep=comma] {Data/hop_comparision_pokec.csv};
	\addplot+[mark=10-pointed star,
	mark size=2.5pt,
	dashed,
	error bars/.cd,
	y fixed,
	y dir=both,
	y explicit
	] table [x={x}, y={Age_NFVR_SVM}, col sep=comma] {Data/hop_comparision_pokec.csv};
	\end{axis}
	\end{tikzpicture}\hspace{-0.1cm}%
	\linebreak	
	\begin{tikzpicture}
	\begin{axis}[title={},
	compat=newest,
	xlabel={{\small $h$}},
	ylabel={Accuracy (\%)}, ylabel shift={-3pt},xtick={},
	height=0.3\columnwidth, width=0.3\columnwidth, grid=major,
	ymin=0, ymax=100,
	xtick={1,2,3},
	legend style={font=\tiny,draw=none,fill=none},
	legend entries={KNN,NB,DT,SVM},
	legend columns = -1,
	legend to name=commonlegend_mape_comparison
	]
	\node[text width=1cm] at (2.5,20) {public};
	\addplot+[
	mark size=2.5pt,
	smooth,
	error bars/.cd,
	y fixed,
	y dir=both,
	y explicit,
	] table [x={x}, y={Public_NNFVR_KNN}, col sep=comma] {Data/hop_comparision_pokec.csv};
	\addplot+[mark=triangle*,
	mark size=2.5pt,
	dashed,
	error bars/.cd,
	y fixed,
	y dir=both,
	y explicit
	] table [x={x}, y={Public_NNFVR_NB}, col sep=comma] {Data/hop_comparision_pokec.csv};
	\addplot+[mark=halfcircle*,
	mark size=2.5pt,
	dashed,
	error bars/.cd,
	y fixed,
	y dir=both,
	y explicit
	] table [x={x}, y={Public_NNFVR_DT}, col sep=comma] {Data/hop_comparision_pokec.csv};
	\addplot+[mark=10-pointed star,
	mark size=2.5pt,
	dashed,
	error bars/.cd,
	y fixed,
	y dir=both,
	y explicit
	] table [x={x}, y={Public_NNFVR_SVM}, col sep=comma] {Data/hop_comparision_pokec.csv};
	\end{axis}
	\end{tikzpicture}\hspace{-0.1cm}%
	\begin{tikzpicture}
	\begin{axis}[title={},
	compat=newest,
	xlabel={{\small $h$}},
	ylabel={},
	height=0.3\columnwidth, width=0.3\columnwidth, grid=major,
	ymin=0, ymax=100,
	xtick={1,2,3},
	yticklabels = {},
	legend style={font=\tiny,draw=none,fill=none},
	legend entries={},
	]
	\node[text width=1cm] at (2.5,20) {gender};
	\addplot+[
	mark size=2.5pt,
	smooth,
	error bars/.cd,
	y fixed,
	y dir=both,
	y explicit
	] table [x={x}, y={Gender_NNFVR_KNN}, col sep=comma] {Data/hop_comparision_pokec.csv};
	\addplot+[mark=triangle*,
	mark size=2.5pt,
	dashed,
	error bars/.cd,
	y fixed,
	y dir=both,
	y explicit
	] table [x={x}, y={Gender_NNFVR_NB}, col sep=comma] {Data/hop_comparision_pokec.csv};
	\addplot+[mark=halfcircle*,
	mark size=2.5pt,
	dashed,
	error bars/.cd,
	y fixed,
	y dir=both,
	y explicit
	] table [x={x}, y={Gender_NNFVR_DT}, col sep=comma] {Data/hop_comparision_pokec.csv};
	\addplot+[mark=10-pointed star,
	mark size=2.5pt,
	dashed,
	error bars/.cd,
	y fixed,
	y dir=both,
	y explicit
	] table [x={x}, y={Gender_NNFVR_SVM}, col sep=comma] {Data/hop_comparision_pokec.csv};
	\end{axis}
	\end{tikzpicture}\hspace{-0.1cm}%
	\begin{tikzpicture}
	\begin{axis}[title={},
	compat=newest,
	xlabel={{\small $h$}},
	ylabel={},xtick={},
	height=0.3\columnwidth, width=0.3\columnwidth, grid=major,
	ymin=0, ymax=100,
	xtick={1,2,3},
	yticklabels = {},
	legend style={font=\tiny,draw=none,fill=none},
	legend entries={},
	]
	\node[text width=1cm] at (2.5,65) {age};
	\addplot+[
	mark size=2.5pt,
	smooth,
	error bars/.cd,
	y fixed,
	y dir=both,
	y explicit
	] table [x={x}, y={Age_NNFVR_KNN}, col sep=comma] {Data/hop_comparision_pokec.csv};
	\addplot+[mark=triangle*,
	mark size=2.5pt,
	dashed,
	error bars/.cd,
	y fixed,
	y dir=both,
	y explicit
	] table [x={x}, y={Age_NNFVR_NB}, col sep=comma] {Data/hop_comparision_pokec.csv};
	\addplot+[mark=halfcircle*,
	mark size=2.5pt,
	dashed,
	error bars/.cd,
	y fixed,
	y dir=both,
	y explicit
	] table [x={x}, y={Age_NNFVR_DT}, col sep=comma] {Data/hop_comparision_pokec.csv};
	\addplot+[mark=10-pointed star,
	mark size=2.5pt,
	dashed,
	error bars/.cd,
	y fixed,
	y dir=both,
	y explicit
	] table [x={x}, y={Age_NNFVR_SVM}, col sep=comma] {Data/hop_comparision_pokec.csv};
	\end{axis}
	\end{tikzpicture}\hspace{-0.1cm}%
	\\	\ref{commonlegend_mape_comparison}
	\caption{$h $-hop effect on \textsc{n-fvr} (top) and \textsc{nn-fvr} (bottom) method using different classifiers for different attributes of Pokec dataset. Figures are best seen in color.}
	\label{NFVR_pokec_h_hop_comparision}
\end{figure}
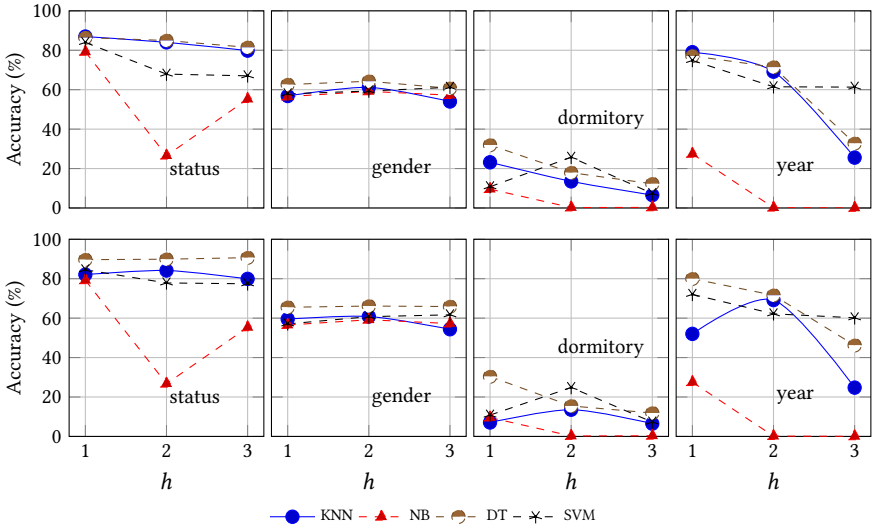
\begin{figure}[t]
	\centering
	\footnotesize
	\begin{tikzpicture}
	\begin{axis}[title={},
	compat=newest,
	ylabel={Accuracy (\%)}, ylabel shift={-3pt},
	ytick = {0,20,40,60,80,100},
	height=0.3\columnwidth, width=0.3\columnwidth, grid=major,
	ymin=0, ymax=100,
	xtick={},
	xticklabels = {},
	legend style={font=\tiny,draw=none,fill=none},
	legend entries={KNN,NB,DT,SVM},
	legend columns = -1,
	legend to name=commonlegend_mape_comparison
	]
	\node[text width=1cm] at (2.5,20) {status};
	\addplot+[
	mark size=2.5pt,
	smooth,
	error bars/.cd,
	y fixed,
	y dir=both,
	y explicit,
	] table [x={x}, y={Status_NFVR_KNN}, col sep=comma] {Data/hop_comparision_unc.csv};
	\addplot+[mark=triangle*,
	mark size=2.5pt,
	dashed,
	error bars/.cd,
	y fixed,
	y dir=both,
	y explicit
	] table [x={x}, y={Status_NFVR_NB}, col sep=comma] {Data/hop_comparision_unc.csv};
	\addplot+[mark=halfcircle*,
	mark size=2.5pt,
	dashed,
	error bars/.cd,
	y fixed,
	y dir=both,
	y explicit
	] table [x={x}, y={Status_NFVR_DT}, col sep=comma] {Data/hop_comparision_unc.csv};
	\addplot+[mark=10-pointed star,
	mark size=2.5pt,
	dashed,
	error bars/.cd,
	y fixed,
	y dir=both,
	y explicit
	] table [x={x}, y={Status_NFVR_SVM}, col sep=comma] {Data/hop_comparision_unc.csv};
	\end{axis}
	\end{tikzpicture}\hspace{-0.1cm}%
	\begin{tikzpicture}
	\begin{axis}[title={},
	compat=newest,
	ylabel={},
	height=0.3\columnwidth, width=0.3\columnwidth, grid=major,
	ymin=0, ymax=100,
	xtick={},
	xticklabels = {},
	yticklabels = {},
	legend style={font=\tiny,draw=none,fill=none},
	legend entries={},
	]
	\node[text width=1cm] at (2.5,20) {gender};
	\addplot+[
	mark size=2.5pt,
	smooth,
	error bars/.cd,
	y fixed,
	y dir=both,
	y explicit
	] table [x={x}, y={Gender_NFVR_KNN}, col sep=comma] {Data/hop_comparision_unc.csv};
	\addplot+[mark=triangle*,
	mark size=2.5pt,
	dashed,
	error bars/.cd,
	y fixed,
	y dir=both,
	y explicit
	] table [x={x}, y={Gender_NFVR_NB}, col sep=comma] {Data/hop_comparision_unc.csv};
	\addplot+[mark=halfcircle*,
	mark size=2.5pt,
	dashed,
	error bars/.cd,
	y fixed,
	y dir=both,
	y explicit
	] table [x={x}, y={Gender_NFVR_DT}, col sep=comma] {Data/hop_comparision_unc.csv};
	\addplot+[mark=10-pointed star,
	mark size=2.5pt,
	dashed,
	error bars/.cd,
	y fixed,
	y dir=both,
	y explicit
	] table [x={x}, y={Gender_NFVR_SVM}, col sep=comma] {Data/hop_comparision_unc.csv};
	\end{axis}
	\end{tikzpicture}\hspace{-0.1cm}%
	\begin{tikzpicture}
	\begin{axis}[title={},
	compat=newest,
	ylabel={},xtick={},
	height=0.3\columnwidth, width=0.3\columnwidth, grid=major,
	ymin=0, ymax=100,
	xtick={},
	xticklabels = {},
	yticklabels = {},
	legend style={font=\tiny,draw=none,fill=none},
	legend entries={},
	]
	\node[text width=1cm] at (2.3,45) {dormitory};
	\addplot+[
	mark size=2.5pt,
	smooth,
	error bars/.cd,
	y fixed,
	y dir=both,
	y explicit
	] table [x={x}, y={Dorm_NFVR_KNN}, col sep=comma] {Data/hop_comparision_unc.csv};
	\addplot+[mark=triangle*,
	mark size=2.5pt,
	dashed,
	error bars/.cd,
	y fixed,
	y dir=both,
	y explicit
	] table [x={x}, y={Dorm_NFVR_NB}, col sep=comma] {Data/hop_comparision_unc.csv};
	\addplot+[mark=halfcircle*,
	mark size=2.5pt,
	dashed,
	error bars/.cd,
	y fixed,
	y dir=both,
	y explicit
	] table [x={x}, y={Dorm_NFVR_DT}, col sep=comma] {Data/hop_comparision_unc.csv};
	\addplot+[mark=10-pointed star,
	mark size=2.5pt,
	dashed,
	error bars/.cd,
	y fixed,
	y dir=both,
	y explicit
	] table [x={x}, y={Dorm_NFVR_SVM}, col sep=comma] {Data/hop_comparision_unc.csv};
	\end{axis}
	\end{tikzpicture}\hspace{-0.1cm}%
	\begin{tikzpicture}
	\begin{axis}[title={},
	compat=newest,
	ylabel={},xtick={},
	height=0.3\columnwidth, width=0.3\columnwidth, grid=major,
	ymin=0, ymax=100,
	xtick={},
	xticklabels = {},
	yticklabels = {},
	legend style={font=\tiny,draw=none,fill=none},
	legend entries={},
	]
	\node[text width=1cm] at (2.5,20) {year};
	\addplot+[
	mark size=2.5pt,
	smooth,
	error bars/.cd,
	y fixed,
	y dir=both,
	y explicit
	] table [x={x}, y={Year_NFVR_KNN}, col sep=comma] {Data/hop_comparision_unc.csv};
	\addplot+[mark=triangle*,
	mark size=2.5pt,
	dashed,
	error bars/.cd,
	y fixed,
	y dir=both,
	y explicit
	] table [x={x}, y={Year_NFVR_NB}, col sep=comma] {Data/hop_comparision_unc.csv};
	\addplot+[mark=halfcircle*,
	mark size=2.5pt,
	dashed,
	error bars/.cd,
	y fixed,
	y dir=both,
	y explicit
	] table [x={x}, y={Year_NFVR_DT}, col sep=comma] {Data/hop_comparision_unc.csv};
	\addplot+[mark=10-pointed star,
	mark size=2.5pt,
	dashed,
	error bars/.cd,
	y fixed,
	y dir=both,
	y explicit
	] table [x={x}, y={Year_NFVR_SVM}, col sep=comma] {Data/hop_comparision_unc.csv};
	\end{axis}
	\end{tikzpicture}\hspace{-0.1cm}%
	\linebreak
	\begin{tikzpicture}
	\begin{axis}[title={},
	compat=newest,
	xlabel={{\small $h$}},
	ylabel={Accuracy (\%)}, ylabel shift={-3pt},xtick={},
	height=0.3\columnwidth, width=0.3\columnwidth, grid=major,
	ymin=0, ymax=100,
	xtick={1,2,3},
	legend style={font=\tiny,draw=none,fill=none},
	legend entries={KNN,NB,DT,SVM},
	legend columns = -1,
	legend to name=commonlegend_mape_comparison
	]
	\node[text width=1cm] at (2.5,20) {status};
	\addplot+[
	mark size=2.5pt,
	smooth,
	error bars/.cd,
	y fixed,
	y dir=both,
	y explicit,
	] table [x={x}, y={Status_NNFVR_KNN}, col sep=comma] {Data/hop_comparision_unc.csv};
	\addplot+[mark=triangle*,
	mark size=2.5pt,
	dashed,
	error bars/.cd,
	y fixed,
	y dir=both,
	y explicit
	] table [x={x}, y={Status_NNFVR_NB}, col sep=comma] {Data/hop_comparision_unc.csv};
	\addplot+[mark=halfcircle*,
	mark size=2.5pt,
	dashed,
	error bars/.cd,
	y fixed,
	y dir=both,
	y explicit
	] table [x={x}, y={Status_NNFVR_DT}, col sep=comma] {Data/hop_comparision_unc.csv};
	\addplot+[mark=10-pointed star,
	mark size=2.5pt,
	dashed,
	error bars/.cd,
	y fixed,
	y dir=both,
	y explicit
	] table [x={x}, y={Status_NNFVR_SVM}, col sep=comma] {Data/hop_comparision_unc.csv};
	\end{axis}
	\end{tikzpicture}\hspace{-0.1cm}%
	\begin{tikzpicture}
	\begin{axis}[title={},
	compat=newest,
	xlabel={{\small $h$}},
	ylabel={},xtick={},
	height=0.3\columnwidth, width=0.3\columnwidth, grid=major,
	ymin=0, ymax=100,
	xtick={1,2,3},
	yticklabels = {},
	legend style={font=\tiny,draw=none,fill=none},
	legend entries={},
	]
	\node[text width=1cm] at (2.5,20) {gender};
	\addplot+[
	mark size=2.5pt,
	smooth,
	error bars/.cd,
	y fixed,
	y dir=both,
	y explicit
	] table [x={x}, y={Gender_NNFVR_KNN}, col sep=comma] {Data/hop_comparision_unc.csv};
	\addplot+[mark=triangle*,
	mark size=2.5pt,
	dashed,
	error bars/.cd,
	y fixed,
	y dir=both,
	y explicit
	] table [x={x}, y={Gender_NNFVR_NB}, col sep=comma] {Data/hop_comparision_unc.csv};
	\addplot+[mark=halfcircle*,
	mark size=2.5pt,
	dashed,
	error bars/.cd,
	y fixed,
	y dir=both,
	y explicit
	] table [x={x}, y={Gender_NNFVR_DT}, col sep=comma] {Data/hop_comparision_unc.csv};
	\addplot+[mark=10-pointed star,
	mark size=2.5pt,
	dashed,
	error bars/.cd,
	y fixed,
	y dir=both,
	y explicit
	] table [x={x}, y={Gender_NNFVR_SVM}, col sep=comma] {Data/hop_comparision_unc.csv};
	\end{axis}
	\end{tikzpicture}\hspace{-0.1cm}%
	\begin{tikzpicture}
	\begin{axis}[title={},
	compat=newest,
	xlabel={{\small $h$}},
	ylabel={},xtick={},
	height=0.3\columnwidth, width=0.3\columnwidth, grid=major,
	ymin=0, ymax=100,
	xtick={1,2,3},
	yticklabels = {},
	legend style={font=\tiny,draw=none,fill=none},
	legend entries={},
	]
	\node[text width=1cm] at (2.3,45) {dormitory};
	\addplot+[
	mark size=2.5pt,
	smooth,
	error bars/.cd,
	y fixed,
	y dir=both,
	y explicit
	] table [x={x}, y={Dorm_NNFVR_KNN}, col sep=comma] {Data/hop_comparision_unc.csv};
	\addplot+[mark=triangle*,
	mark size=2.5pt,
	dashed,
	error bars/.cd,
	y fixed,
	y dir=both,
	y explicit
	] table [x={x}, y={Dorm_NNFVR_NB}, col sep=comma] {Data/hop_comparision_unc.csv};
	\addplot+[mark=halfcircle*,
	mark size=2.5pt,
	dashed,
	error bars/.cd,
	y fixed,
	y dir=both,
	y explicit
	] table [x={x}, y={Dorm_NNFVR_DT}, col sep=comma] {Data/hop_comparision_unc.csv};
	\addplot+[mark=10-pointed star,
	mark size=2.5pt,
	dashed,
	error bars/.cd,
	y fixed,
	y dir=both,
	y explicit
	] table [x={x}, y={Dorm_NNFVR_SVM}, col sep=comma] {Data/hop_comparision_unc.csv};
	\end{axis}
	\end{tikzpicture}\hspace{-0.1cm}%
	\begin{tikzpicture}
	\begin{axis}[title={},
	compat=newest,
	xlabel={{\small $h$}},
	ylabel={},xtick={},
	height=0.3\columnwidth, width=0.3\columnwidth, grid=major,
	ymin=0, ymax=100,
	xtick={1,2,3},
	yticklabels = {},
	legend style={font=\tiny,draw=none,fill=none},
	legend entries={},
	]
	\node[text width=1cm] at (2.5,20) {year};
	\addplot+[
	mark size=2.5pt,
	smooth,
	error bars/.cd,
	y fixed,
	y dir=both,
	y explicit
	] table [x={x}, y={Year_NNFVR_KNN}, col sep=comma] {Data/hop_comparision_unc.csv};
	\addplot+[mark=triangle*,
	mark size=2.5pt,
	dashed,
	error bars/.cd,
	y fixed,
	y dir=both,
	y explicit
	] table [x={x}, y={Year_NNFVR_NB}, col sep=comma] {Data/hop_comparision_unc.csv};
	\addplot+[mark=halfcircle*,
	mark size=2.5pt,
	dashed,
	error bars/.cd,
	y fixed,
	y dir=both,
	y explicit
	] table [x={x}, y={Year_NNFVR_DT}, col sep=comma] {Data/hop_comparision_unc.csv};
	\addplot+[mark=10-pointed star,
	mark size=2.5pt,
	dashed,
	error bars/.cd,
	y fixed,
	y dir=both,
	y explicit
	] table [x={x}, y={Year_NNFVR_SVM}, col sep=comma] {Data/hop_comparision_unc.csv};
	\end{axis}
	\end{tikzpicture}\hspace{-0.1cm}%
	\\	\ref{commonlegend_mape_comparison}
	\caption{$h $-hop effect on \textsc{n-fvr} (top) and \textsc{nn-fvr} (bottom) method using different classifiers for different attributes of UNC dataset. Figures are best seen in color.}
	\label{NFVR_unc_h_hop_comparision}
\end{figure}
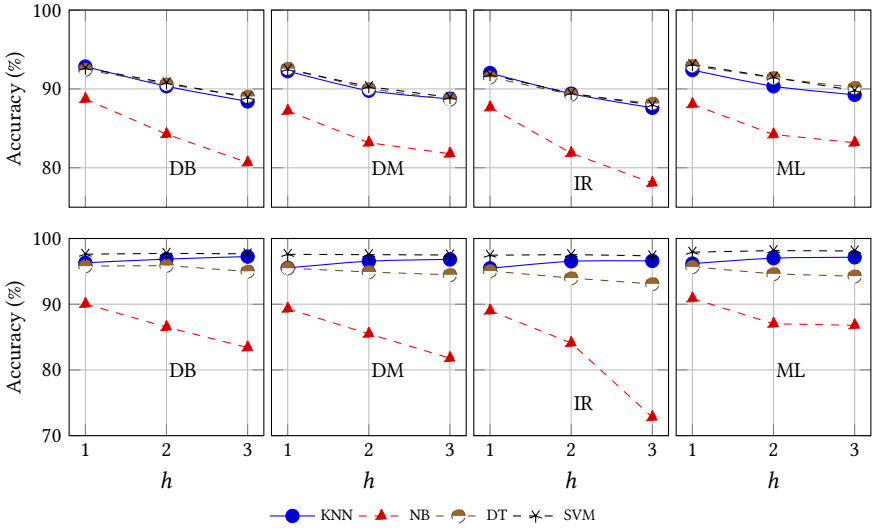
\begin{figure}[t]
	\centering
	\footnotesize
	\begin{tikzpicture}
	\begin{axis}[title={},
	compat=newest,
	ylabel={Accuracy (\%)}, ylabel shift={-3pt},xtick={},
	height=0.3\columnwidth, width=0.3\columnwidth, grid=major,
	ymin=75, ymax=100,
	xtick={1,2,3},
	xticklabels = {},
	legend style={font=\tiny,draw=none,fill=none},
	legend entries={KNN,NB,DT,SVM},
	legend columns = -1,
	legend to name=commonlegend_mape_comparison
	]
	\node[text width=1cm] at (2.5,80) {DB};
	\addplot+[
	mark size=2.5pt,
	smooth,
	error bars/.cd,
	y fixed,
	y dir=both,
	y explicit,
	] table [x={x}, y={DB_NFVR_KNN}, col sep=comma] {Data/hop_comparision4area.csv};
	\addplot+[mark=triangle*,
	mark size=2.5pt,
	dashed,
	error bars/.cd,
	y fixed,
	y dir=both,
	y explicit
	] table [x={x}, y={DB_NFVR_NB}, col sep=comma] {Data/hop_comparision4area.csv};
	\addplot+[mark=halfcircle*,
	mark size=2.5pt,
	dashed,
	error bars/.cd,
	y fixed,
	y dir=both,
	y explicit
	] table [x={x}, y={DB_NFVR_DT}, col sep=comma] {Data/hop_comparision4area.csv};
	\addplot+[mark=10-pointed star,
	mark size=2.5pt,
	dashed,
	error bars/.cd,
	y fixed,
	y dir=both,
	y explicit
	] table [x={x}, y={DB_NFVR_SVM}, col sep=comma] {Data/hop_comparision4area.csv};
	\end{axis}
	\end{tikzpicture}\hspace{-0.1cm}%
	\begin{tikzpicture}
	\begin{axis}[title={},
	compat=newest,
	ylabel={},
	height=0.3\columnwidth, width=0.3\columnwidth, grid=major,
	ymin=75, ymax=100,
	xtick={1,2,3},
	xticklabels = {},
	yticklabels = {},
	legend style={font=\tiny,draw=none,fill=none},
	legend entries={},
	]
	\node[text width=1cm] at (2.5,80) {DM};
	\addplot+[
	mark size=2.5pt,
	smooth,
	error bars/.cd,
	y fixed,
	y dir=both,
	y explicit
	] table [x={x}, y={DM_NFVR_KNN}, col sep=comma] {Data/hop_comparision4area.csv};
	\addplot+[mark=triangle*,
	mark size=2.5pt,
	dashed,
	error bars/.cd,
	y fixed,
	y dir=both,
	y explicit
	] table [x={x}, y={DM_NFVR_NB}, col sep=comma] {Data/hop_comparision4area.csv};
	\addplot+[mark=halfcircle*,
	mark size=2.5pt,
	dashed,
	error bars/.cd,
	y fixed,
	y dir=both,
	y explicit
	] table [x={x}, y={DM_NFVR_DT}, col sep=comma] {Data/hop_comparision4area.csv};
	\addplot+[mark=10-pointed star,
	mark size=2.5pt,
	dashed,
	error bars/.cd,
	y fixed,
	y dir=both,
	y explicit
	] table [x={x}, y={DM_NFVR_SVM}, col sep=comma] {Data/hop_comparision4area.csv};
	\end{axis}
	\end{tikzpicture}\hspace{-0.1cm}%
	\begin{tikzpicture}
	\begin{axis}[title={},
	compat=newest,
	ylabel={},xtick={},
	height=0.3\columnwidth, width=0.3\columnwidth, grid=major,
	ymin=75, ymax=100,
	xtick={1,2,3},
	xticklabels = {},
	yticklabels = {},
	legend style={font=\tiny,draw=none,fill=none},
	legend entries={},
	]
	\node[text width=1cm] at (2.5,78) {IR};
	\addplot+[
	mark size=2.5pt,
	smooth,
	error bars/.cd,
	y fixed,
	y dir=both,
	y explicit
	] table [x={x}, y={IR_NFVR_KNN}, col sep=comma] {Data/hop_comparision4area.csv};
	\addplot+[mark=triangle*,
	mark size=2.5pt,
	dashed,
	error bars/.cd,
	y fixed,
	y dir=both,
	y explicit
	] table [x={x}, y={IR_NFVR_NB}, col sep=comma] {Data/hop_comparision4area.csv};
	\addplot+[mark=halfcircle*,
	mark size=2.5pt,
	dashed,
	error bars/.cd,
	y fixed,
	y dir=both,
	y explicit
	] table [x={x}, y={IR_NFVR_DT}, col sep=comma] {Data/hop_comparision4area.csv};
	\addplot+[mark=10-pointed star,
	mark size=2.5pt,
	dashed,
	error bars/.cd,
	y fixed,
	y dir=both,
	y explicit
	] table [x={x}, y={IR_NFVR_SVM}, col sep=comma] {Data/hop_comparision4area.csv};
	\end{axis}
	\end{tikzpicture}\hspace{-0.1cm}%
	\begin{tikzpicture}
	\begin{axis}[title={},
	compat=newest,
	ylabel={},xtick={},
	height=0.3\columnwidth, width=0.3\columnwidth, grid=major,
	ymin=75, ymax=100,
	xtick={1,2,3},
	xticklabels = {},
	yticklabels = {},
	legend style={font=\tiny,draw=none,fill=none},
	legend entries={},
	]
	\node[text width=1cm] at (2.5,80) {ML};
	\addplot+[
	mark size=2.5pt,
	smooth,
	error bars/.cd,
	y fixed,
	y dir=both,
	y explicit
	] table [x={x}, y={ML_NFVR_KNN}, col sep=comma] {Data/hop_comparision4area.csv};
	\addplot+[mark=triangle*,
	mark size=2.5pt,
	dashed,
	error bars/.cd,
	y fixed,
	y dir=both,
	y explicit
	] table [x={x}, y={ML_NFVR_NB}, col sep=comma] {Data/hop_comparision4area.csv};
	\addplot+[mark=halfcircle*,
	mark size=2.5pt,
	dashed,
	error bars/.cd,
	y fixed,
	y dir=both,
	y explicit
	] table [x={x}, y={ML_NFVR_DT}, col sep=comma] {Data/hop_comparision4area.csv};
	\addplot+[mark=10-pointed star,
	mark size=2.5pt,
	dashed,
	error bars/.cd,
	y fixed,
	y dir=both,
	y explicit
	] table [x={x}, y={ML_NFVR_SVM}, col sep=comma] {Data/hop_comparision4area.csv};
	\end{axis}
	\end{tikzpicture}\hspace{-0.1cm}%
	\linebreak
	\begin{tikzpicture}
	\begin{axis}[title={},
	compat=newest,
	xlabel={{\small $h$}},
	ylabel={Accuracy (\%)}, ylabel shift={-3pt},xtick={},
	height=0.3\columnwidth, width=0.3\columnwidth, grid=major,
	ymin=70, ymax=100,
	xtick={1,2,3},
	legend style={font=\tiny,draw=none,fill=none},
	legend entries={KNN,NB,DT,SVM},
	legend columns = -1,
	legend to name=commonlegend_mape_comparison
	]
	\node[text width=1cm] at (2.5,80) {DB};
	\addplot+[
	mark size=2.5pt,
	smooth,
	error bars/.cd,
	y fixed,
	y dir=both,
	y explicit,
	] table [x={x}, y={DB_NNFVR_KNN}, col sep=comma] {Data/hop_comparision4area.csv};
	\addplot+[mark=triangle*,
	mark size=2.5pt,
	dashed,
	error bars/.cd,
	y fixed,
	y dir=both,
	y explicit
	] table [x={x}, y={DB_NNFVR_NB}, col sep=comma] {Data/hop_comparision4area.csv};
	\addplot+[mark=halfcircle*,
	mark size=2.5pt,
	dashed,
	error bars/.cd,
	y fixed,
	y dir=both,
	y explicit
	] table [x={x}, y={DB_NNFVR_DT}, col sep=comma] {Data/hop_comparision4area.csv};
	\addplot+[mark=10-pointed star,
	mark size=2.5pt,
	dashed,
	error bars/.cd,
	y fixed,
	y dir=both,
	y explicit
	] table [x={x}, y={DB_NNFVR_SVM}, col sep=comma] {Data/hop_comparision4area.csv};
	\end{axis}
	\end{tikzpicture}\hspace{-0.1cm}%
	\begin{tikzpicture}
	\begin{axis}[title={},
	compat=newest,
	xlabel={{\small $h$}},
	ylabel={},xtick={},
	height=0.3\columnwidth, width=0.3\columnwidth, grid=major,
	ymin=70, ymax=100,
	xtick={1,2,3},
	yticklabels = {},
	legend style={font=\tiny,draw=none,fill=none},
	legend entries={},
	]
	\node[text width=1cm] at (2.5,80) {DM};
	\addplot+[
	mark size=2.5pt,
	smooth,
	error bars/.cd,
	y fixed,
	y dir=both,
	y explicit
	] table [x={x}, y={DM_NNFVR_KNN}, col sep=comma] {Data/hop_comparision4area.csv};
	\addplot+[mark=triangle*,
	mark size=2.5pt,
	dashed,
	error bars/.cd,
	y fixed,
	y dir=both,
	y explicit
	] table [x={x}, y={DM_NNFVR_NB}, col sep=comma] {Data/hop_comparision4area.csv};
	\addplot+[mark=halfcircle*,
	mark size=2.5pt,
	dashed,
	error bars/.cd,
	y fixed,
	y dir=both,
	y explicit
	] table [x={x}, y={DM_NNFVR_DT}, col sep=comma] {Data/hop_comparision4area.csv};
	\addplot+[mark=10-pointed star,
	mark size=2.5pt,
	dashed,
	error bars/.cd,
	y fixed,
	y dir=both,
	y explicit
	] table [x={x}, y={DM_NNFVR_SVM}, col sep=comma] {Data/hop_comparision4area.csv};
	\end{axis}
	\end{tikzpicture}\hspace{-0.1cm}%
	\begin{tikzpicture}
	\begin{axis}[title={},
	compat=newest,
	xlabel={{\small $h$}},
	ylabel={},xtick={},
	height=0.3\columnwidth, width=0.3\columnwidth, grid=major,
	ymin=70, ymax=100,
	xtick={1,2,3},
	yticklabels = {},
	legend style={font=\tiny,draw=none,fill=none},
	legend entries={},
	]
	\node[text width=1cm] at (2.5,75) {IR};
	\addplot+[
	mark size=2.5pt,
	smooth,
	error bars/.cd,
	y fixed,
	y dir=both,
	y explicit
	] table [x={x}, y={IR_NNFVR_KNN}, col sep=comma] {Data/hop_comparision4area.csv};
	\addplot+[mark=triangle*,
	mark size=2.5pt,
	dashed,
	error bars/.cd,
	y fixed,
	y dir=both,
	y explicit
	] table [x={x}, y={IR_NNFVR_NB}, col sep=comma] {Data/hop_comparision4area.csv};
	\addplot+[mark=halfcircle*,
	mark size=2.5pt,
	dashed,
	error bars/.cd,
	y fixed,
	y dir=both,
	y explicit
	] table [x={x}, y={IR_NNFVR_DT}, col sep=comma] {Data/hop_comparision4area.csv};
	\addplot+[mark=10-pointed star,
	mark size=2.5pt,
	dashed,
	error bars/.cd,
	y fixed,
	y dir=both,
	y explicit
	] table [x={x}, y={IR_NNFVR_SVM}, col sep=comma] {Data/hop_comparision4area.csv};
	\end{axis}
	\end{tikzpicture}\hspace{-0.1cm}%
	\begin{tikzpicture}
	\begin{axis}[title={},
	compat=newest,
	xlabel={{\small $h$}},
	ylabel={},xtick={},
	height=0.3\columnwidth, width=0.3\columnwidth, grid=major,
	ymin=70, ymax=100,
	xtick={1,2,3},
	yticklabels = {},
	legend style={font=\tiny,draw=none,fill=none},
	legend entries={},
	]
	\node[text width=1cm] at (2.5,80) {ML};
	\addplot+[
	mark size=2.5pt,
	smooth,
	error bars/.cd,
	y fixed,
	y dir=both,
	y explicit
	] table [x={x}, y={ML_NNFVR_KNN}, col sep=comma] {Data/hop_comparision4area.csv};
	\addplot+[mark=triangle*,
	mark size=2.5pt,
	dashed,
	error bars/.cd,
	y fixed,
	y dir=both,
	y explicit
	] table [x={x}, y={ML_NNFVR_NB}, col sep=comma] {Data/hop_comparision4area.csv};
	\addplot+[mark=halfcircle*,
	mark size=2.5pt,
	dashed,
	error bars/.cd,
	y fixed,
	y dir=both,
	y explicit
	] table [x={x}, y={ML_NNFVR_DT}, col sep=comma] {Data/hop_comparision4area.csv};
	\addplot+[mark=10-pointed star,
	mark size=2.5pt,
	dashed,
	error bars/.cd,
	y fixed,
	y dir=both,
	y explicit
	] table [x={x}, y={ML_NNFVR_SVM}, col sep=comma] {Data/hop_comparision4area.csv};
	\end{axis}
	\end{tikzpicture}\hspace{-0.1cm}%
	\\	\ref{commonlegend_mape_comparison}
	\caption{$h $-hop effect on \textsc{n-fvr} (top) and \textsc{nn-fvr} (bottom) method using different classifiers for different attributes of 4area dataset. Figures are best seen in color.}
	\label{NFVR_4area_h_hop_comparision}
\end{figure}

In regards to UNC dataset, the results for $h$-hop are presented in Figure~\ref{NFVR_unc_h_hop_comparision}. In majority of the cases, we observe decrease in performance as we increase the value of $h$. This is true for both the \textsc{n-fvr} and the \textsc{nn-fvr} based experiments.

Turning now to 4area dataset, the results are presented in Figure~\ref{NFVR_4area_h_hop_comparision}. With respect to \textsc{n-fvr} approach, the predictive performance of all classifier tend to decrease while using the $h$-hop value greater than $1$. While in case of \textsc{nn-fvr}, only \textsc{knn} shows a slight improvement for $h$-hop values of $1$ and $2$ while other classifiers show the decrease in performance for any value of $h$-hop greater than $1$. These extra results on the value of $h$-hop showed that the choice of $h$-hop value depends on the type of classifier used. The value of this parameter cannot be generalized across datasets and attributes. However, $h$-hop$=1$ is the most optimal overall.

Similar behavior for $h$-hop neighborhood is observed in case of UChicago, Temple, Haverford, and Mississippi datasets. Their results are not shown because of space constraint.

\end{document}